\documentclass[twoside]{article}

\usepackage[accepted]{aistats2026}

\usepackage[english]{babel}

\usepackage{amsmath}
\usepackage{amssymb}
\usepackage{graphicx}
\usepackage{float}
\usepackage[colorlinks=true, allcolors=blue]{hyperref}
\usepackage{cleveref}
\usepackage{natbib}
\usepackage{booktabs}
\usepackage{multirow}
\usepackage[ruled,vlined]{algorithm2e}
\newcommand{\dd}{\mathop{}\!\mathrm{d}}
\newcommand{\norm}[1]{\left\lVert #1 \right\rVert}

\newcommand{\E}[2]{\mathbb{E}_{#1 \sim #2}}
\newcommand{\KL}[2]{D_{\text{KL}} \left[ #1 \ \vert \vert \ #2 \right]}
\newcommand{\Score}[2]{\displaystyle S^{#1_{#2}}}
\newcommand{\cent}[1]{\mathcal{H} \left( #1 \right)}
\newcommand{\Eqref}[1]{\hyperref[#1]{Eq.~(\ref*{#1})}}
\newcommand{\Secref}[1]{\hyperref[#1]{\S~\ref*{#1}}}
\newcommand{\Figref}[1]{\hyperref[#1]{Figure~\ref*{#1}}}
\newcommand{\Algoref}[1]{\hyperref[#1]{Algorithm~\ref*{#1}}}

\begin{document}


\twocolumn[

\aistatstitle{TENDE: Transfer Entropy Neural Diffusion Estimation}

\runningauthor{Galeano Mu\~{n}oz, Bounoua, Franzese, Michiardi, Filippone}

\aistatsauthor{
Simon Pedro Galeano Mu\~{n}oz \And 
Mustapha Bounoua \And  
Giulio Franzese \AND
Pietro Michiardi \And
Maurizio Filippone
}

\vspace{-1.1cm}

\aistatsaddress{ KAUST, Saudi Arabia \And  EURECOM, France \And EURECOM, France \AND EURECOM, France \And KAUST, Saudi Arabia } 

]

\begin{abstract}
Transfer entropy is a fundamental measure for quantifying directed information flow in time series, with applications spanning neuroscience, finance, and complex systems analysis. However, existing estimation methods suffer from the curse of dimensionality, require restrictive distributional assumptions, or need exponentially large datasets for reliable convergence. We address these limitations in the literature by proposing TENDE (Transfer Entropy Neural Diffusion Estimation), a novel approach that leverages score-based diffusion models to estimate transfer entropy through conditional mutual information. By learning score functions of the relevant conditional distributions, TENDE provides flexible, scalable estimation while making minimal assumptions about the underlying data-generating process. We demonstrate superior accuracy and robustness compared to existing neural estimators and other state-of-the-art approaches across synthetic benchmarks and real data.
\end{abstract}

\section{Introduction}
\label{sec:introduction}

Estimating dependencies between variables is a fundamental problem in Statistics and Machine Learning. For time series, this becomes particularly important in applications in neuroscience \citep{parente2021modelling,el2025methods,wang2025time}, where researchers analyze information flow between brain regions, and in finance \citep{patton2012review, gong2022asymmetric, caserini2022effective}, where understanding relationships between assets is crucial for risk assessment.
The challenge is to quantify these dependencies without assuming specific functional relationships between time series, while making minimal distributional assumptions. Transfer entropy (TE), introduced by \citet{schreiber2000measuring}, addresses this by measuring directed information flow between time series through conditional mutual information (CMI). However, the high-dimensional nature of the problem, considering both current values and historical lags, makes reliable estimation difficult. 

Existing methods face significant limitations. Traditional approaches based on k-nearest neighbors \citep{lindner2011trentool} suffer from the curse of dimensionality. Recent neural estimators using variational bounds \citep{zhang2019itene} can require exponentially large datasets for convergence \citep{mcallester2020formal}, while copula-based methods \citep{redondo2023measuring} or the use of entropy arguments \citep{kornai2025agm} impose restrictive assumptions.
Recent advances in score-based diffusion models \citep{song2020score} offer a promising solution. These models excel at learning complex probability distributions by estimating score functions, and accurate density estimation is sufficient for computing information-theoretic measures. Building on connections between diffusion models and KL divergence estimation \citep{franzese2023minde}, we can leverage these advances for transfer entropy estimation.

In this work, we propose TENDE (Transfer Entropy Neural Diffusion Estimation), which uses score-based diffusion models to estimate transfer entropy. Our approach is flexible and scalable, and makes minimal distributional assumptions while providing accurate estimates even in high-dimensional settings.

The paper is organized as follows: \Secref{sec:background} introduces the fundamental concepts of transfer entropy and its formal definition. \Secref{sec:related} reviews related work on estimation methods, and \Secref{sec:methods} presents our diffusion-based estimator. \Secref{sec:experiment:synthetic} provides a comparative analysis against KNN, copula, cross-entropy, and Donsker-Varadhan based approaches. \Secref{sec:experiment:real} demonstrates the method on the Santa Fe B time series dataset to illustrate its practical applicability. \Secref{sec:conclusions} concludes discussing future directions.

\section{Background}
\label{sec:background}
\subsection{Mutual Information and Conditional Mutual Information}
\label{sec:MI}
Capturing the dependence between random variables is a recurrent problem in several applications of Statistics and Machine Learning. The Mutual Information (MI) is an attractive measure of dependence when the relation between the variables is unknown and possibly nonlinear. The MI is defined as follows: let $X \in \mathbb{R}^{N_\mathrm{x}}$ and $Y \in \mathbb{R}^{N_\mathrm{y}}$ be random variables with joint probability density $p_{X, Y}$ and marginal densities $p_X \text{ and } p_Y$ respectively\footnote{MI can be defined also for variables without densities and in more generic spaces, but for the purpose of this work the restriction considered here is sufficient.}, the mutual information between $X$ and $Y$ is given by 
\begin{equation}
    I(X; Y) = \KL{p_{X, Y}}{p_{X} \ p_{Y}},
\end{equation}
where $\KL{p}{q}$ denotes the Kullback–Leibler (KL) Divergence between the distributions $p$ and $q$ and is defined as
\begin{equation}
     \KL{p}{q} =  \E{x}{p} \left[ \log \left( \frac{p(x)}{q(x)} \right) \right].
\end{equation}
It is worth recalling that the MI between $X$ and $Y$ equals zero if and only if $p_{X, Y} = p_{X} \ p_{Y}$, that is, if and only if $X$ and $Y$ are independent random variables. While MI has been widely employed in diverse domains, it only captures unconditional pairwise dependencies. In many applications, however, the relationship between two variables may be driven by a set of other variables. To address this, MI naturally extends to its conditional form, the conditional mutual information, which quantifies the dependence between two random variables $X$ and $Y$ given a third variable $Z$. Formally, CMI is defined as follows:
\begin{equation}
    \label{eq:CMI}
    I(X; Y \vert Z) =  \E{z}{p_Z} \left[ I(X_z; Y_z) \right].
\end{equation}
Here $X_z$ and $Y_z$ denote the random variables $X \vert Z = z$ and $Y \vert Z = z$, thus \Eqref{eq:CMI} represents the average MI between $X$ and $Y$ where $Z$ is known, that is, the mean KL divergence between  $p_{X_z, Y_z}$ and $p_{X_z} \ p_{Y_z}$ where $p_{X_z, Y_z}$, $p_{X_z}$, and $p_{Y_z}$ represent the joint density of $X \text{ and } Y$ and the marginal densities of $X \text{ and } Y$ conditioned on $Z = z$ respectively. Analogously $p_Z$ is the marginal density of the random variable $Z$.

This perspective is crucial in scenarios where the apparent association between $X$ and $Y$ may be entirely driven by their joint dependence on $Z$, rather than reflecting a direct relationship. By conditioning on $Z$, CMI provides a principled way to disentangle direct from indirect dependencies, offering a more refined characterization of the underlying dependence structure. Such considerations are particularly important in complex systems where interactions among variables are often mediated through latent or observed confounders.

Although MI and CMI are measures of general dependence between random, neither can capture the directionality of the dependence since we have $I(X; Y) = I(Y; X)$ and $I(X; Y \vert Z ) = I(Y; X \vert Z )$; the equalities can be easily seen from the symmetric form in which the joint and marginal distributions appear in the KL divergence.  
In many applications such as the ones described in  \citet{baccala2001partial,kayser2009comparison,wang2022different,cirstian2023objective}, it is highly desirable to identify not only whether two variables are dependent but also the direction of dependence, as this could provide insight into the underlying mechanisms that govern the system at hand. Without accounting for directionality, analyses may overlook critical asymmetries in the flow of information that determine how complex systems evolve.
\subsection{Transfer Entropy}
\label{sec:te_def}
To solve this issue \citet{schreiber2000measuring} developed the concept of transfer entropy. Let $\{X_t \}$ and $\{ Y_t \}$ denote $N_\mathrm{x}$-dimensional and $N_\mathrm{y}$-dimensional time series, respectively. Define 
$$
\begin{cases}
\begin{aligned}
\mathbf{Y}_{t-\ell} &= \left[ Y_{t - 1}, \ \dots, \ Y_{t - \ell} \right] \\    
\mathbf{X}_{t-k} &= \left[ X_{t - 1}, \ \dots, \ X_{t - k} \right],
\end{aligned}    
\end{cases}
$$
for some natural numbers $\ell, k $. Thus, the TE from $\{X_t \}$ to $\{ Y_t \}$ is given by 
\begin{equation}\label{eq:trans_entr}
\mathrm{TE}_{X \to Y}(k,\ell) = I \left ( Y_t; \mathbf{X}_{t - k} \vert \mathbf{Y}_{t - \ell} \right).    
\end{equation}
TE quantifies how much $Y_t$ depends on the past of $\{X_t\}$ once its past is already known. If $Y_t$ is independent of $\mathbf{X}_{t - k}$ once $\mathbf{Y}_{t - \ell}$ is observed, then $TE(X \rightarrow Y; k, \ell) = 0$. Hence, a positive transfer entropy indicates that the past of $\{X_t\}$ contains unique predictive information about $Y_t$ that is not already present in its own history.
 It can be observed from the definition TE that it is not symmetric, this is because in general $$I \left ( X_t; \mathbf{Y}_{t - k} \vert \mathbf{X}_{t - \ell} \right) \neq I \left ( Y_t; \mathbf{X}_{t - k} \vert \mathbf{Y}_{t - \ell} \right)\text{.}$$ 
Transfer entropy has thus become a widely used tool for analyzing directed dependencies in time.
However, its practical application is often limited by challenges related to reliable estimation, particularly in finite-sample and high-dimensional settings \citep{Zhao2020KNN, Gao2018KNN}.
\section{Related work}
\label{sec:related}
There are several proposals in the literature on how to estimate the TE between two time series. The first class of proposed methods for this matter, such as the work by \citet{lindner2011trentool} is based on the use of $k$-nearest neighbors, leveraging the entropy representation of TE. These estimators are inspired by the methodology described by \citet{frenzel2007partial}, which uses the approach by \citet{kozachenko1987sample} to estimate the entropy terms. Although these classical methods remained popular for their ease of use, theoretical and experimental results suggest that they suffer from the curse of dimensionality, as discussed in \citet{Zhao2020KNN, Gao2018KNN}. 

More recently, copulas were used to estimate TE using the fact that MI can be represented as the copula entropy \citep{ma2011mutual}. \citet{redondo2023measuring} exploit the ability of copulas to decouple marginal effects from the dependence structure, thereby improving the robustness and interpretability in TE estimation. Nevertheless, the simplifying assumption commonly employed in vine copula decompositions \citep{bedford2002vines} to mitigate the curse of dimensionality does not always hold in practice, as demonstrated by \citet{derumigny2020kendall} and \citet{gijbels2021omnibus}. A more comprehensive discussion of this issue is provided by \citet{nagler2025simplified}.

In parallel, neural estimators have been proposed to overcome the limitations of both $k$-nearest neighbors and copula-based methods. These approaches leverage the expressive power of neural networks to model complex, nonlinear dependencies between time series without requiring explicit assumptions about the underlying distributions. Among recent proposals, there are two main concepts that are used as the building blocks for the estimation of TE. On the one hand, approaches such as \citep{zhang2019itene, luxembourg2025treet} take advantage of the Donsker-Varadhan variational lower bound on the KL divergence; however, the arguments provided by \citet{mcallester2020formal} imply that methods using this lower bound as means to compute TE require exponentially large datasets. On the other hand, the proposals of \citet{garg2022estimating, shalev2022neural}, and \citet{kornai2025agm} use cross-entropy arguments to compute the TE, following the suggestion that methods using upper bounds on entropies will not suffer convergence issues of variational approaches. Despite overcoming the limitations of variational methods, \citet{garg2022estimating} and \citet{shalev2022neural} use categorical distributions as means to compute the TE.
Even though \citet{kornai2025agm} overcame this limitation by avoiding categorical distributions in favor of a parametric estimation of the conditionals, the need to choose a parametric form represents a limitation. We also note the related line of work on neural estimation of directed information for sequential settings \citep{tsur2023neural}, and approaches based on sliced mutual information \citep{goldfeld2021sliced, tsur2023maxsliced} that address the curse of dimensionality through lower-dimensional projections.
\section{Methods}
\label{sec:methods}
\subsection{General overview of score-based KL divergence estimation}
\label{sec:overview_scores}
Recent developments in generative modeling \citep{song2020score} and information-theoretic learning have opened new avenues for TE estimation. In particular, score-based diffusion models provide a principled mechanism to approximate data distributions through the estimation of their score functions, thereby enabling flexible modeling of high-dimensional systems. Parallel to this, advances in mutual information estimation \citep{franzese2023minde, KongBS23} have improved the accuracy and scalability of this task in less restrictive scenarios. A natural extension is to integrate these two approaches, leveraging the expressive power of diffusion models for distributional representation, while employing modern mutual information and entropy estimators to compute CMI as the building block to quantify directional dependencies.

Recall that $X$ denotes a $N_\mathrm{x}$-dimensional random variable with probability distribution $p_X$. Under certain regularity conditions, \citet{hyvarinen2005estimation} showed that it is possible to associate the density $p_X$ with the score function $\Score{p}{X}$, where for a generic distribution $p_X$ we denote $\Score{p}{X}(x) := \nabla \log \left( p_{X}(x) \right)$, with derivatives taken with respect to $x$. In addition, it is possible to construct a diffusion process $\{X_t\}_{t \in [0, T]}$ such that $X_0 \sim p_X$ and $X_T \sim p_{X_T}$ where $p_{X_T}$ is a distribution such that there is a tractable way to sample efficiently from it. This diffusion process is modeled as the solution of the following stochastic differential equation:
\begin{equation}
    \label{eq:diffusion}
    \begin{cases}
        \begin{aligned}
            dX_t &= f_t X_t dt + g_t dW_t \\
            X_0 &\sim p_X,
        \end{aligned}
    \end{cases}
\end{equation}
with given continuous functions $f_t \leq 0, \ g_t \geq 0$ for each $t \in [0, T]$, and $dW_t$ is a Brownian motion. The random variable $X_t$ is associated with its density $p_{X_t}$ and therefore with the time-varying score $\Score{p}{X_t}(x)$.

One of the results by \citet{bounoua2024s} (see also \citet{franzese2023minde}) states that if there is another probability density $q_X$, serving as a reference distribution, for which $q_{X_t}$ is generated by the same diffusion process described in \Eqref{eq:diffusion}, then the KL divergence between $p_X$ and $q_X$ can be expressed as
\begin{equation}
    \label{eq:kld_diff}
    \begin{aligned}
    \int_0^{T} \frac{g_t^2}{2}  & \E{x}{p_{X_t}} \left[\left \lVert \Score{p}{X_t}(x) - \Score{q}{X_t}(x) \right \rVert^2 \right] dt \\
    &+ D_{KL} \left[p_{X_T} \vert \vert q_{X_T} \right],
    \end{aligned}
\end{equation}
where $\Vert \cdot \rVert$ denotes the standard Euclidean norm in $\mathbb{R}^{N_\mathrm{x}}$.

This result is a remarkable way to link KL divergence with diffusion processes, given the knowledge on the score functions of $p_{X_t}$ and $q_{X_t}$. Nonetheless, the availability of such objects is out of reach in practical applications, and that is why this work instead considers parametric approximations of scores. Thus, for a generic distribution $p$, its score $\Score{p}{X}(x)$ is approximated by a neural network $\Score{p}{X}(x; \theta^{\star})$ where $\theta^\star$ is obtained by minimizing the loss of denoising score matching \citep{vincent2011connection}. Thus, as stated in \citet{song2020score} for the case of the time-varying score, $\theta^\star$ is obtained by minimizing
\begin{equation}
    \label{eq:obj}
    \int_0^T \E{x}{p} \E{\tilde{x} \vert x}{p_{0t}} \left[\left \lVert \Score{p}{X_t}(\tilde{x}; \theta) - \Score{p}{0t}(\tilde{x} \vert x) \right \rVert^2 \right] dt,
\end{equation}
where $p_{0t}(\cdot \vert x)$ denotes the transition density of $X_t$ conditioned on $X_0 = x$, and $\Score{p}{0t}(\tilde{x} \vert x) = \nabla_{\tilde{x}} \log p_{0t}(\tilde{x} \vert x)$ is the corresponding score function evaluated at $\tilde{x}$. The marginal score $\Score{p}{X_t}$ at diffusion time $t$ is the quantity being approximated by the neural network. The term inside the integral in \Eqref{eq:obj} is equivalent to
\begin{equation}
    \int p(x) p_{0t}(\tilde{x} \vert x) \left \lVert \Score{p}{X_t}(\tilde{x}; \theta) - \Score{p}{0t}(\tilde{x} \vert x) \right \rVert^2 d\tilde{x} dx.
\end{equation}
Following the work of \citet{franzese2023minde}, we adopt the quantity $e(p, q)$ as an estimator of the KL divergence between $p$ and $q$, with
%
\begin{align}
\label{eq:kld_estimator}
e&(p, q) = \nonumber \\  
&\int_0^{T} \frac{g_t^2}{2}  \E{x}{p_t} \left[\left \lVert \Score{p}{X_t}(x; \theta_1^\star) - \Score{q}{X_t}(x; \theta_2^\star) \right \rVert^2 \right] dt.
\end{align}
%
This is simply the first term of \Eqref{eq:kld_diff}, where parametric scores are used instead of the true score functions. Under the assumption that the learned scores are sufficiently accurate, the terminal KL divergence $D_{KL}[p_{X_T} \vert \vert q_{X_T}]$ becomes negligible for large $T$, and thus \citep{franzese2023minde}
\begin{equation*}
    \begin{aligned}
        e(p, q) \simeq D_{KL} [p \vert \vert q].
    \end{aligned}
\end{equation*}
A detailed discussion of the approximation error, decomposed into the score estimation error and the terminal divergence, is provided in \Secref{sec:approx_error}.
\subsection{Score-based entropy estimation}
\label{sec:score_est}
We now turn our attention to the estimation of entropy using score functions. For this, consider $X$ as previously defined in \Secref{sec:MI}, the entropy is defined as ${H}(X) = \E{x}{p_X} \left[- \log \ p_X(x) \right]$, thus it is possible to relate the entropy of a random variable with the KL divergence in the following manner. Let $\varphi_{\sigma}(\cdot)$ denote the density of a $N_\mathrm{x}$-dimensional centered Gaussian random variable with covariance $\sigma^2 \mathbf{I}_{N_\mathrm{x}}$, then the entropy of $X$ can be written as 
\begin{equation}
\label{eq:ent_kld}
\begin{aligned}
{H}&(X) = \\
&\frac{N_\mathrm{x}}{2} \log(2 \pi \sigma^2) + \E{x}{p_X} \left[ \frac{\lVert x \rVert^2}{2 \sigma^2} \right] - D_{KL}[p_X \vert \vert \varphi_\sigma].
\end{aligned}
\end{equation}
Thus, it can be shown that the entropy of $X$ can be estimated as 
\begin{equation}
\label{eq:ent_kld_hat}
    \begin{aligned}
        {H}(X; \sigma) &\simeq \frac{N_\mathrm{x}}{2} \log \left( 2 \pi \sigma^2 \right) + \E{x}{p_X} \left[ \frac{\lVert x \rVert^2}{2 \sigma^2} \right] \\
        & - e(p_X, \varphi_\sigma) - \frac{N_\mathrm{x}}{2} \left( \log(\chi_T) - 1 + \frac{1}{\chi_T} \right).
    \end{aligned}
\end{equation} 
Where for $t \in [0, T]$, $\chi_t = \left( k_t^2 \sigma^2 + k_t^2 \int_0^t \frac{g_s^2}{k_s^2} ds \right)$ with $k_t = \exp\left \{\int_0^t f_s ds \right \}$. The derivations of \Eqref{eq:ent_kld} and \Eqref{eq:ent_kld_hat} can be found in \Secref{sec:entr_gauss}. 
\subsection{Score-based conditional mutual information and transfer entropy estimation}
\label{sec:estimatecmi} 
In this work, we are interested in the estimation of TE, which is formulated in terms of CMI. For ease of exposition, we provide estimators of the CMI and then state how to use such estimators to compute TE between two time series. Consider random variables $X \in \mathbb{R}^{N_\mathrm{x}}$, $Y \in \mathbb{R}^{N_\mathrm{y}}$, and $Z \in \mathbb{R}^{N_\mathrm{z}}$. The main result in \citet{franzese2023minde} provides an accurate way to estimate the KL divergence between two densities $p$ and $q$ utilizing diffusion models, so quantities such as MI or entropies can be estimated since they can be represented in terms of KL divergences. The notation for random variables, conditional random variables, and their respective densities remains analogous to the notation used in \Secref{sec:background}. With this in mind, we take advantage of the following expressions that are equivalent to CMI
\begin{align}
    I(X;Y\vert Z) \label{eq:te_kld}&= \E{[y, z]}{p_{Y, Z}} \left[ \KL{p_{X_{y, z}}}{p_{X_z}} \right], \\
    \label{eq:te_ents}&= \cent{X \vert Z} - \cent{X \vert Y, Z}, \\
    \label{eq:mis}&= I(X; [Y, Z]) - I(X; Z),
\end{align}
where $\cent{X \vert Z} = \E{z}{p_z} \left[ H(X_z)\right]$, the definition of $\cent{X \vert Z, Y}$ is analogous.

Using the estimator $e(\cdot, \cdot)$ from \Eqref{eq:kld_estimator} to approximate each KL divergence term, we obtain the following CMI estimators 
\begin{equation}
\label{eq:cmihat1}
\begin{aligned}
    \E{[y, z]}{p_{Y, Z}} &\left[\KL{p_{X_{y, z}}}{p_{X_z}} \right] \simeq \\ 
    &\E{[y, z]}{p_{Y, Z}} \left[ e(p_{X_{y, z}}, p_{X_z}) \right],
\end{aligned}
\end{equation}
\begin{equation}
\label{eq:cmihat2}
    \begin{aligned}
    &\cent{X \vert Z} - \cent{X \vert Y, Z} \simeq \\
    &\E{[y, z]}{p_{Y, Z}} \left[ e(p_{X_{y, z}}, \varphi_\sigma)\right] - \E{z}{p_{Z}} \left[ e(p_{X_{z}}, \varphi_\sigma)\right],
    \end{aligned}
\end{equation}
\begin{equation}
\label{eq:cmihat3}
    \begin{aligned}
        &I(X; [Y, Z]) - I(X; Z) \simeq \\
        &\E{[y, z]}{p_{Y, Z}} \left[ e(p_{X_{y, z}}, p_{X})\right] - \E{z}{p_{Z}} \left[ e(p_{X_{z}}, p_{X}) \right].
    \end{aligned}
\end{equation}
It is worth mentioning that it is possible to perturb the conditional entropy terms in \Eqref{eq:cmihat2} by adding and subtracting $e(p_X, \varphi_\sigma)$ appropriately, leading to individual estimators for $I(X; [Y, Z])$ and $I(X; Z)$. As a result, we also propose the following estimator for CMI
\begin{equation}
\label{eq:cmihat4}
    CMI(X; Y \vert Z) \simeq \hat{I}(X; [Y, Z]) - \hat{I}(X; Z),
\end{equation}
with 
\begin{equation*}
\begin{aligned}
\hat{I}(X; [Y, Z]) = \E{[y, z]}{p_{Y, Z}} \left[ e(p_{X_{y, z}}, p_{X})\right] - e(p_X, \varphi_\sigma),
\end{aligned}
\end{equation*}
and
\begin{equation*}
    \begin{aligned}
        \hat{I}(X; Z) = \E{z}{p_{Z}} \left[ e(p_{X_{z}}, p_{X}) \right] - e(p_X, \varphi_\sigma).
    \end{aligned}
\end{equation*}
Among the proposed estimators, \Eqref{eq:cmihat1} is generally preferable as it is guaranteed to be non-negative, since it directly estimates a KL divergence. The estimators in \Eqref{eq:cmihat2}--\Eqref{eq:cmihat4} are valuable when the individual components (e.g., conditional entropies or mutual informations) are of independent interest; however, as difference-based estimators, they may be more susceptible to error propagation. Derivations of the estimators are available in \Secref{sec:derivs_ests}.
\subsubsection{TE estimation}
\label{sec:te_estimation_description}
Let $\{X_t\}_{t=1}^T$ be the source series with dimensionality $N_\mathrm{x}$, and let $\{Y_t\}_{t=1}^T$ be the target series with dimensionality $N_\mathrm{y}$. Choose source and target lags $k,\ell \in \mathbb{N}$. For each time index $t$ with $t > \max(k,\ell)$, a sample is constructed as follows. The future target is given by $Y := Y_t \in \mathbb{R}^{N_\mathrm{y}}$. The past of the source is represented as $X := [X_{t-1}, X_{t-2}, \dots, X_{t-k}] \in \mathbb{R}^{kN_\mathrm{x}}$. The past of the target, lags as the conditioning set, is represented as $Z := [Y_{t-1}, Y_{t-2}, \dots, Y_{t-\ell}] \in \mathbb{R}^{\ell N_\mathrm{y}}$.

Stacking these triplets for $t = \max(k,\ell) + 1, \dots, T$ produces a dataset 
\begin{equation*}
\{(X^{(i)}, Y^{(i)}, Z^{(i)})\}_{i=1}^{T-\max(k,\ell)},    
\end{equation*}
which can be directly employed for conditional mutual information estimation. When the underlying processes $\{X_t\}$ and $\{Y_t\}$ are jointly stationary, each window follows the same distribution, so sample averages over temporal windows serve as ergodic approximations of the required expectations. By definition, the transfer entropy from $X$ to $Y$ with lags $(k,\ell)$ is then expressed as
\begin{equation*}
\mathrm{TE}_{X \to Y}(k,\ell) = I(Y;X \vert Z).    
\end{equation*}
Once this dataset is constructed, it can be used to train our proposed score-based conditional mutual information estimator and compute the TE.
The way in which $\mathrm{TE}_{Y \to X}(\ell, k)$ can be computed is analogous to what is described above by simply exchanging the roles between $\{X_t\}$ and $\{Y_t\}$.
\subsubsection{Algorithm overview}
In this work, we employ the variance preserving stochastic differential equation as described in \citet{song2020score} to construct the diffusion process. A key practical advantage of the VP formulation is that the transition density $p_{0t}(\cdot \vert x)$ is available in closed form as a Gaussian, so obtaining diffused data at any time $t$ requires only sampling from this known distribution rather than numerically solving the SDE. Leveraging the implementation of \citet{e26040320}, we make use of a single score network that approximates all the score functions required to estimate transfer entropy, amortizing the learning of two or three score functions into a single model. In \Algoref{algo:tende}, the \texttt{conditional} approach groups the estimators in \Eqref{eq:cmihat1} and \Eqref{eq:cmihat2}, which rely only on conditional scores, while the \texttt{joint} approach groups the estimators in \Eqref{eq:cmihat3} and \Eqref{eq:cmihat4}, which additionally require the marginal score. The implementation to estimate the TE in the direction $Y \to X$ is obtained by swapping the roles of $X_t$ and $Y_t$. Regarding the encoding in the third argument of the network, $1$ indicates the variable for which the score is learned, $-1$ denotes that the corresponding input is marginalized out (set to zero), and $0$ indicates that the input is treated as a conditioning signal. Additional details on the network architecture and the amortization procedure are provided in \Secref{sec:algos}.
\begin{algorithm}[h!] 

\DontPrintSemicolon
\SetAlgoLined
\SetNoFillComment
\SetKwInOut{Parameter}{parameter}
\LinesNotNumbered 
\caption{{TENDE}}
\label{algo:tende}
\KwData{$ [X_t , Y_t]$ }  
\Parameter{\texttt{approach} $\in \{\texttt{conditional}, \texttt{joint}\}$, $\sigma$, \texttt{estimator} $\in \{1, 2\}$}

Obtain $Y, X, Z$ as described in \Secref{sec:te_estimation_description} \\
$t^\star \sim \mathcal{U}[0,T]$ \\
\tcp*{diffuse signals to timestep $t^\star$}
$ [Y_{t^\star}, X_{t^\star}, Z_{t^\star}] \gets k_{t^\star}[Y,X, Z]+\left(k^2_{t^\star}\int_0^{t^\star} k^{-2}_sg^2_{s}\dd s\right)^{\frac{1}{2}} [\epsilon_1,\epsilon_2, \epsilon_3]$, with $\epsilon_{1,2, 3} \sim \gamma_1$ \\
\tcp*{Use the score network to compute the required scores}
\uIf{\texttt{approach} = \texttt{conditional}}{
$\Score{p}{Y_{t^{\star}}}_{x, z} \gets net_\theta([Y_{t^\star} , X, Z], t^\star, [1, 0, 0] ) $ \\
$\Score{p}{Y_{t^{\star}}}_{z} \gets net_\theta([Y_{t^\star} , X, Z], t^\star, [1, -1, 0])$ \\
\uIf{\texttt{estimator} $= 1$}{ 
  $\hat{I} \gets  T \frac{g^2_{t^\star} }{2} \norm{\Score{p}{Y_{t^{\star}}}_{x, z} - \Score{p}{Y_{t^{\star}}}_{z}}^2  $ \Eqref{eq:cmihat1}
}
\uElse{ 
    $\chi_{t^\star} \gets  \left(k^2_{t^\star}\sigma^2+k^2_{t^\star}\int_0^{t^\star} k^{-2}_sg^2_{s}\dd s\right) $ \\
    $I_1 \gets \norm{\Score{p}{Y_{t^{\star}}}_{x, z} + \frac{Y_{t^\star} }{\chi_{t^\star}}  }^2$ \\
    $I_2 \gets \norm{\Score{p}{Y_{t^{\star}}}_{z}  +\frac{Y_{t^\star}}{\chi_{t^\star}} }^2$ \\
    $\hat{I}  \gets T \frac{g^2_{t^\star}}{2}   \left[I_1  - I_2 \right ]$ \Eqref{eq:cmihat2}
}

}
\uElse{
$\Score{p}{X_{t^{\star}}}_{y, z} \gets net_\theta([Y_{t^\star} , X, Z], t^\star, [1,0,0]) $ \\
$\Score{p}{X_{t^{\star}}}_{z} \gets net_\theta([Y_{t^\star} , X, Z], t^\star, [1,-1,0]) $ \\
$\Score{p}{X_{t^{\star}}} \gets net_\theta([[Y_{t^\star} , X, Z], t^\star, [1,-1, -1]) $\\

\uIf{\texttt{estimator} $= 1$}{
  $I_1 \gets \norm{\Score{p}{Y_{t^{\star}}}_{x, z} - \Score{p}{Y_{t^{\star}}}}^2$ \\
    $I_2 \gets \norm{\Score{p}{Y_{t^{\star}}}_{z} - \Score{p}{Y_{t^{\star}}}}^2$ \\
  $\hat{I} \gets  T \frac{g^2_{t^\star} }{2} \left(I_1 - I_2\right)$ \Eqref{eq:cmihat3}
}
\uElse{ 
    $\chi_{t^\star} \gets  \left(k^2_{t^\star}\sigma^2+k^2_{t^\star}\int_0^{t^\star} k^{-2}_sg^2_{s}\dd s\right)$  \\
    $I_1 \gets \norm{\Score{p}{Y_{t^{\star}}}_{x, z} + \frac{Y_{t^\star} }{\chi_{t^\star}}  }^2 - \norm{\Score{p}{Y_{t^{\star}}} + \frac{Y_{t^\star} }{\chi_{t^\star}}  }^2$ \\
    $I_2 \gets \norm{\Score{p}{Y_{t^{\star}}}_{z} + \frac{Y_{t^\star} }{\chi_{t^\star}}  }^2 - \norm{\Score{p}{Y_{t^{\star}}} + \frac{Y_{t^\star} }{\chi_{t^\star}}  }^2$ \\
    $\hat{I}  \gets T \frac{g^2_{t^\star}}{2} \left(I_1 - I_2 \right)$ \Eqref{eq:cmihat4}
} 
}
\Return $\hat{I}$
\end{algorithm}
\section{Synthetic benchmark}
\label{sec:experiment:synthetic}

We now evaluate the estimators proposed in \Secref{sec:estimatecmi} using the benchmark by \citet{kornai2025agm} testing our estimators against the methods by \citet{kornai2025agm} (Agm), \citet{steeg_information-theoretic_2013} (Npeet), an adaptation of \citep{ishmael2018mine} (MINE) to compute conditional mutual information as a means of computing TE, the Transformer-based estimator TREET \citep{luxembourg2025treet}, and the conditional independence testing framework implemented in Tigramite \citep{runge2019detecting}.

The empirical validation uses two different types of time series for which the TE is known. The first of these is given by a two-dimensional vector autoregressive process of order $1$ which can be described as follows:
\begin{equation}
\label{eq:lgsys}
\begin{cases}
    \begin{aligned}
        x_t &= b_x x_{t-1} + \lambda y_{t-1} + \varepsilon^{x}_t\\
        y_t &= b_y y_{t-1} + \varepsilon^{y}_t,
    \end{aligned}
\end{cases}
\end{equation}
where both $\varepsilon^{x}_t$ and $\varepsilon^{y}_t$ are independent zero-mean Gaussian innovations with variances $\sigma^2_{x}$ and $\sigma^2_{y}$ respectively. As it can be seen in \Eqref{eq:lgsys}, $y_t$ is independent of the past of $x_t$ so the TE from $X$ to $Y$ is zero. Furthermore, note that $x_t$ depends on the past of $y_t$ so the TE is positive. A closed form for this expression can be found in \citet{edinburgh2021causality}. We refer to this process as \textbf{linear Gaussian system} in the figures.

The second kind of time series is a bivariate process whose realizations are generated according to the following scheme. Let $x_t \sim N(0, 1)$ and $z_t \sim N(0, 1)$ be independent, let $\rho \in (-1, 1)$, and construct $y_t$ as follows:
\begin{equation}
            y_t = 
            \begin{cases}
            z_{t-1}, \ y_{t-1} < \lambda, \\
            \rho x_{t-1} + \sqrt{1 - \rho^2} z_{t-1}, \ y_{t-1} \geq \lambda. 
            \end{cases} 
\end{equation}
Thus, the bivariate system is given by $\left[x_t, y_t\right]$: we refer to this process as \textbf{joint system} in the figures. In this case, the TE from $Y$ to $X$ is null, but as shown by \citet{zhang2019itene}, the TE in the other direction is given by $-\frac{1}{2} (1 - \Phi(\lambda)) \log(1 - \rho^2)$, where $\Phi(\cdot)$ is the cumulative distribution function of a standard Gaussian random variable. 
It can be seen from the processes described above that in both cases the parameter $\lambda$ controls the strength of the dependency measured by the TE between the components of the system.
\subsection{TE estimation benchmark}
\label{sec:estimation_benchmark}
\textbf{Benchmarking.} We consider four different tasks to evaluate the performance of the estimators. For all tasks, each reported result corresponds to the average of estimations over 5 seeds, where for every seed a new dataset is generated and the model is reinitialized and retrained from the ground up. Following the setup by \citet{kornai2025agm}, we use $10000$ samples to estimate the transfer entropy in all tasks except for the sample size benchmark. Moreover, $\lambda$ is fixed to $0$ in the Gaussian system and to $0.5$ in the joint system for the tasks in which $\lambda$ is not varied, while the remaining parameters are kept consistent with those \citep{kornai2025agm}. More experiments can be found in \Secref{sec:addexps}.

\paragraph{Sample size effect.} We focus on computing the transfer entropy for varying sample sizes to analyze how the accuracy of the estimates improves as the number of observations increases. In this case, the different sample sizes considered for both systems are $T = 500, 1000, 5000, 10000$.  

\begin{figure}[h!]
    \centering
    \includegraphics[width=0.47\textwidth]{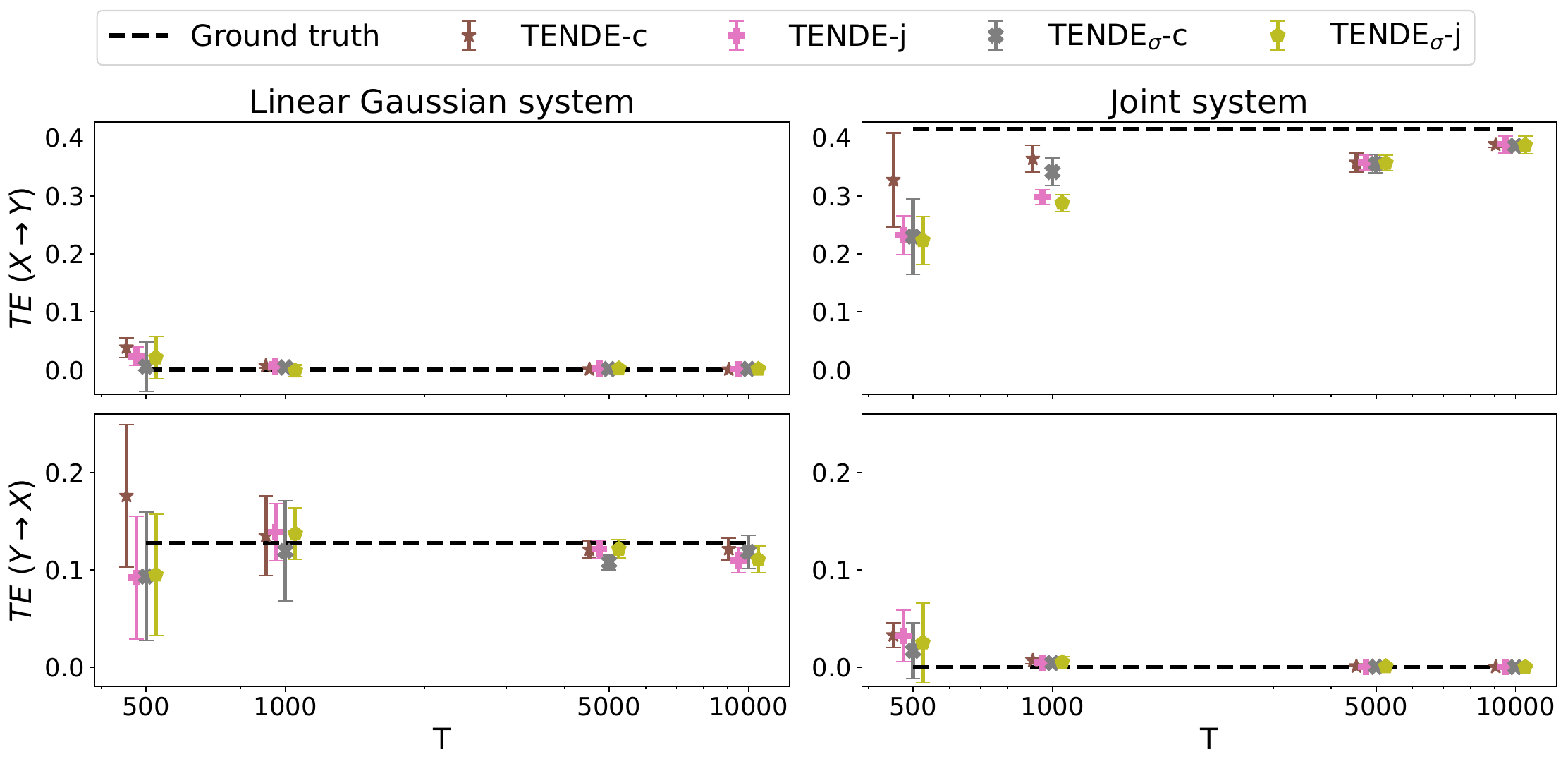}
    \caption{Transfer entropy estimation across sample sizes for linear Gaussian and joint systems.}
    \label{fig:sample_size}
\end{figure}
\paragraph{Consistency.} We examine a two-dimensional system where the parameter $\lambda$ is varied, allowing us to study how changes in coupling strength affect the measured transfer entropy. For this matter, we simulate both systems using nine evenly distributed values of $\lambda$ between $0$ and $1$.
\begin{figure}[h!]
    \centering
    \includegraphics[width=0.47\textwidth]{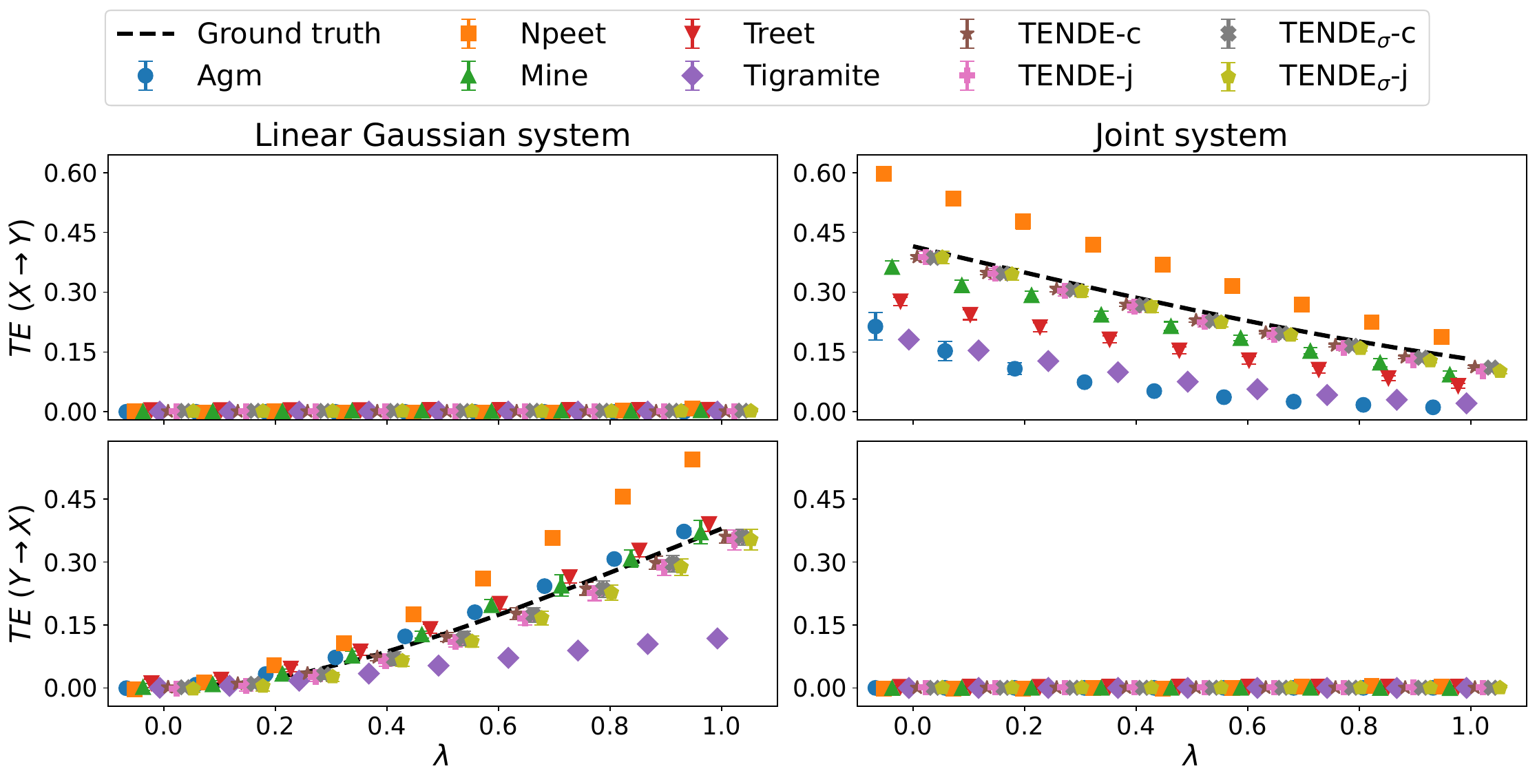}
    \caption{Transfer entropy estimation for varying coupling strength ($\lambda$).}
    \label{fig:simple}
\end{figure}

\paragraph{Redundant stacking.} We stack $d$ redundant dimensions onto both $x_t$ and $y_t$, but which do not contribute to the transfer entropy. More precisely, we consider a $2d$-dimensional time series $\left[\tilde{x}_t, \tilde{y}_t\right]$ with 
\begin{equation}
    \begin{cases}
        \begin{aligned}
            \tilde{x}_t &= \left[x_t, \varepsilon^{x}_{t, 1}, \dots \varepsilon^{x}_{t, d} \right] \\
            \tilde{y}_t &= \left[y_t,  \varepsilon^{y}_{t, 1}, \dots \varepsilon^{y}_{t, d} \right],
        \end{aligned}
    \end{cases}
\end{equation}

where the redundant dimensions $\left( \varepsilon^{x}_{t, i}, \varepsilon^{x}_{t, j} \right)$ are independent Gaussian white noise processes for \newline $1 \leq i, j \leq d$, hence $\mathrm{TE}_{\tilde{X} \to \tilde{Y}}(k,\ell) = \mathrm{TE}_{X \to Y}(k,\ell)$. A proof of this fact is available in \Secref{sec:pf1}

\begin{figure}[t]
    \centering
    \includegraphics[width=0.47\textwidth]{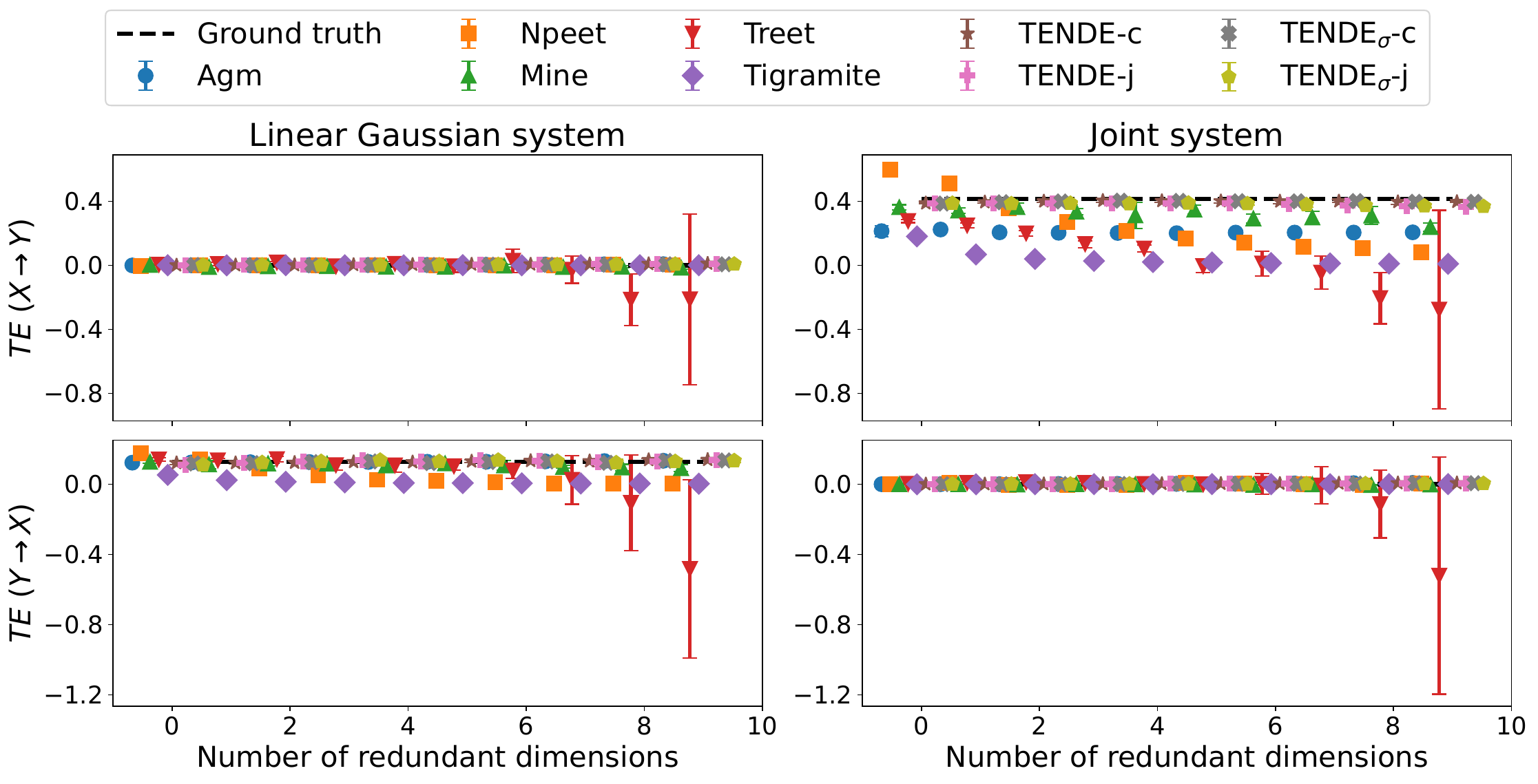}
    \caption{Transfer entropy estimation with added redundant (noise) dimensions.}
    \label{fig:redundant_dimensions}
\end{figure}

\paragraph{Linear stacking.} We consider a scenario in which $d$ replicates of the processes $x_t$ and $y_t$ are stacked in such a way that dependence exists only between corresponding components, making the transfer entropy additive across dimensions. That is, the $2d$-dimensional time series $\left[\tilde{x}_t, \tilde{y}_t\right]$ is given by 
\begin{equation}
    \begin{cases}
        \begin{aligned}
            \tilde{x}_t &= \left[x_{t, 1}, \dots x_{t, d} \right] \\
            \tilde{y}_t &= \left[y_{t, 1}, \dots y_{t, d} \right],
        \end{aligned}
    \end{cases}
\end{equation}
where both collections of processes $\left\{ x_{t, i}\right\}$ and $\left\{ y_{t, i}\right\}$ for $1 \leq i \leq d$ are independent replicates of $x_t$ and $y_t$ respectively, that is, $x_{t, i} \perp y_{t, j}$ for $i \neq j$ and $x_{t, i} \not\perp y_{t, j}$ if $i = j$, thus the transfer entropy between is given by $\mathrm{TE}_{\tilde{X} \to \tilde{Y}}(k,\ell) = \sum_{i = 1}^{d} \mathrm{TE}_{X_i \to Y_i}(k,\ell)$. The details for this fact are provided in \Secref{sec:pf2}.
\begin{figure}[t]
    \centering
    \includegraphics[width=0.47\textwidth]{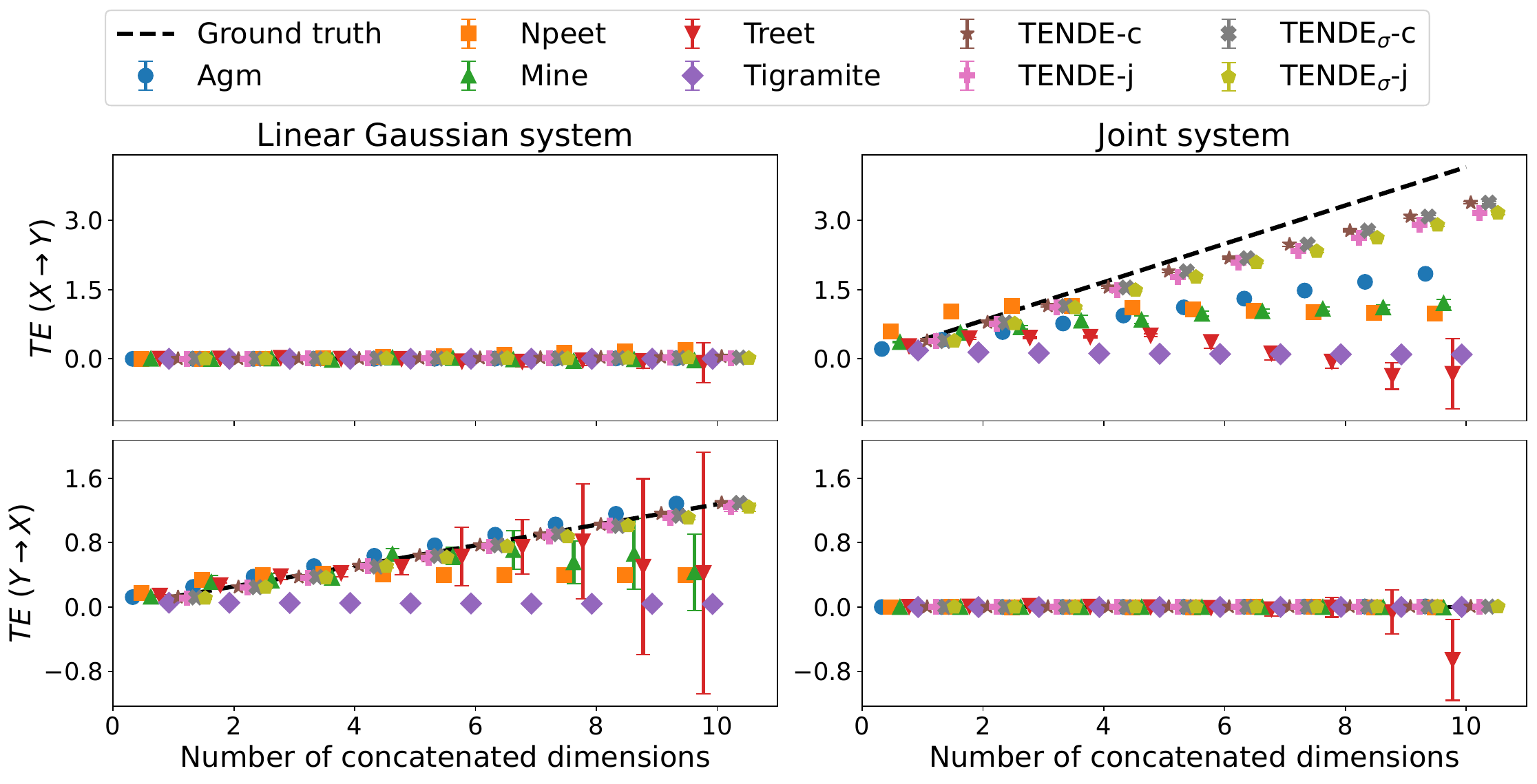}
    \caption{Transfer entropy estimation for linearly stacked systems where multiple independent process copies create additive transfer entropy.}
    \label{fig:linear_stacking}
\end{figure}
\paragraph{Discussion.} The synthetic benchmark results demonstrate TENDE's superior performance across all evaluation scenarios, particularly in high-dimensional settings where traditional methods fail. In the sample size experiments (\Figref{fig:sample_size}), our estimators converge reliably to the ground truth as data increases. When varying the coupling strength (\Figref{fig:simple}), TENDE 
\begin{figure*}[t]
   \centering
   \includegraphics[width=0.19\textwidth]{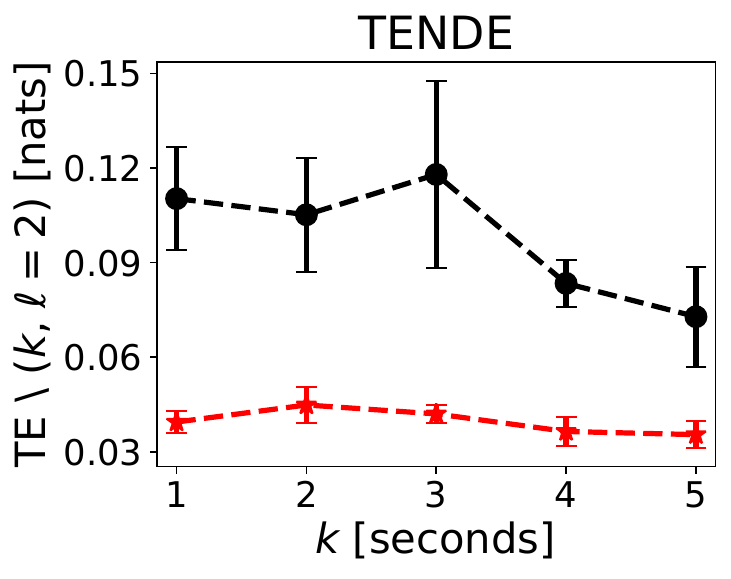} 
   \includegraphics[width=0.19\textwidth]{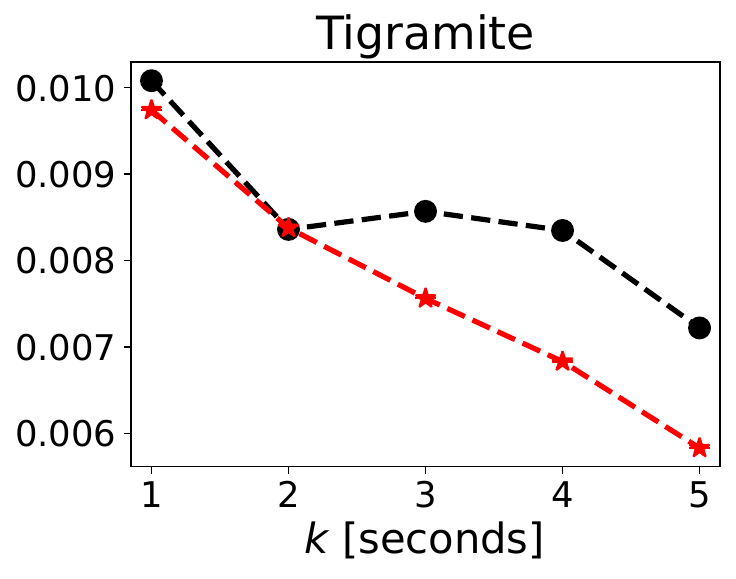} 
   \includegraphics[width=0.19\textwidth]{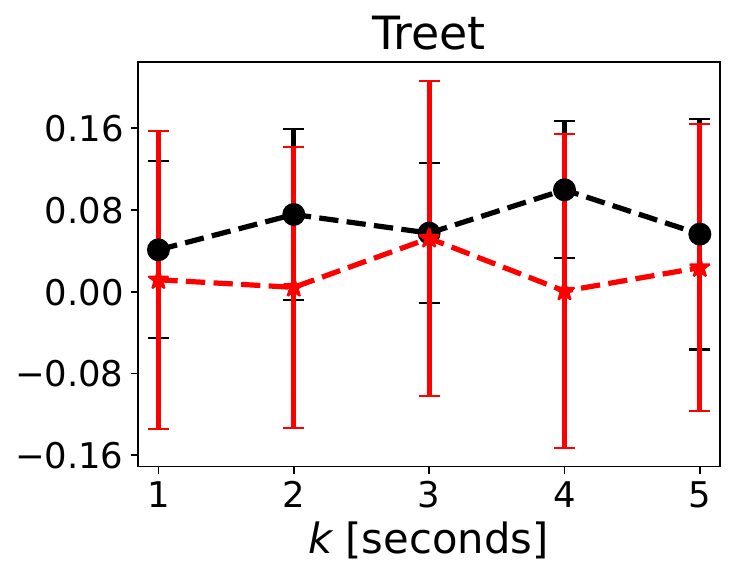} 
   \includegraphics[width=0.19\textwidth]{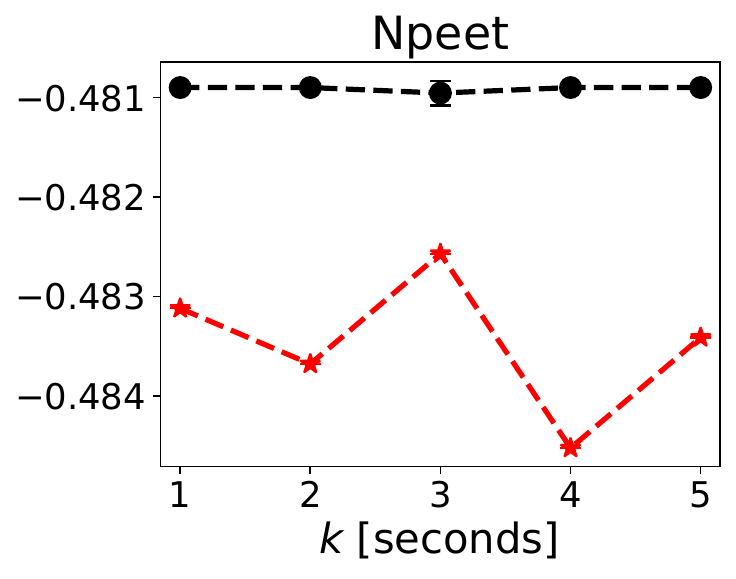} 
   \includegraphics[width=0.19\textwidth]{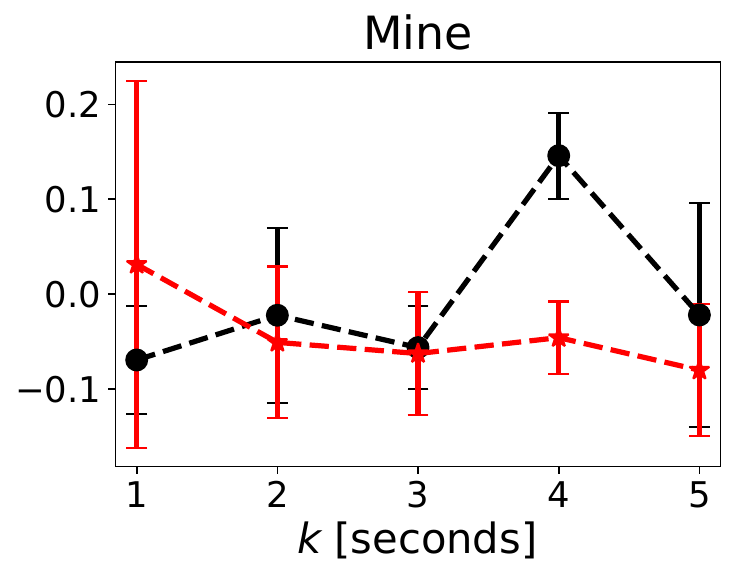}
   \caption{
   Transfer entropy $TE(k,\ell=2)$ between breathing and heart signals as a function of lag $k$ for each estimator. Black denotes the TE from the respiration force to the heart rate, whereas red denotes TE in the other direction. The reported error bars correspond to the standard deviations over 5 seeds.}
   \label{fig:results_real}
\end{figure*}
accurately captures the expected trends, unlike competing estimators that show instability. Under redundant stacking (\Figref{fig:redundant_dimensions}), our approach remains robust to irrelevant noise dimensions, maintaining stable estimates while others degrade sharply; notably, TREET exhibits large variance and produces negative estimates at higher dimensions, highlighting the instability of variational approaches in this regime. 
Tigramite consistently underestimates the transfer entropy, yielding near-zero values. Finally, in the linearly stacked setting (\Figref{fig:linear_stacking}), TENDE scales additively with the number of independent process copies, matching theoretical expectations, whereas both TREET and Tigramite fail to track the growing ground truth. These results highlight that the score-based framework naturally handles complex conditional distributions without restrictive assumptions, contrasting with $k$-nearest neighbor methods that suffer from the curse of dimensionality and variational approaches requiring exponentially large datasets. While AGM performs well under correct parametric assumptions, TENDE achieves comparable or superior performance without such prior knowledge, making it a more robust and practical estimator for real-world applications. Additional results at higher dimensions and with larger sample sizes are reported in \Secref{sec:addexps}.
\section{Real data analysis}
\label{sec:experiment:real}
\begin{figure}[!htb]
   \centering
   \includegraphics[width=0.94\linewidth]{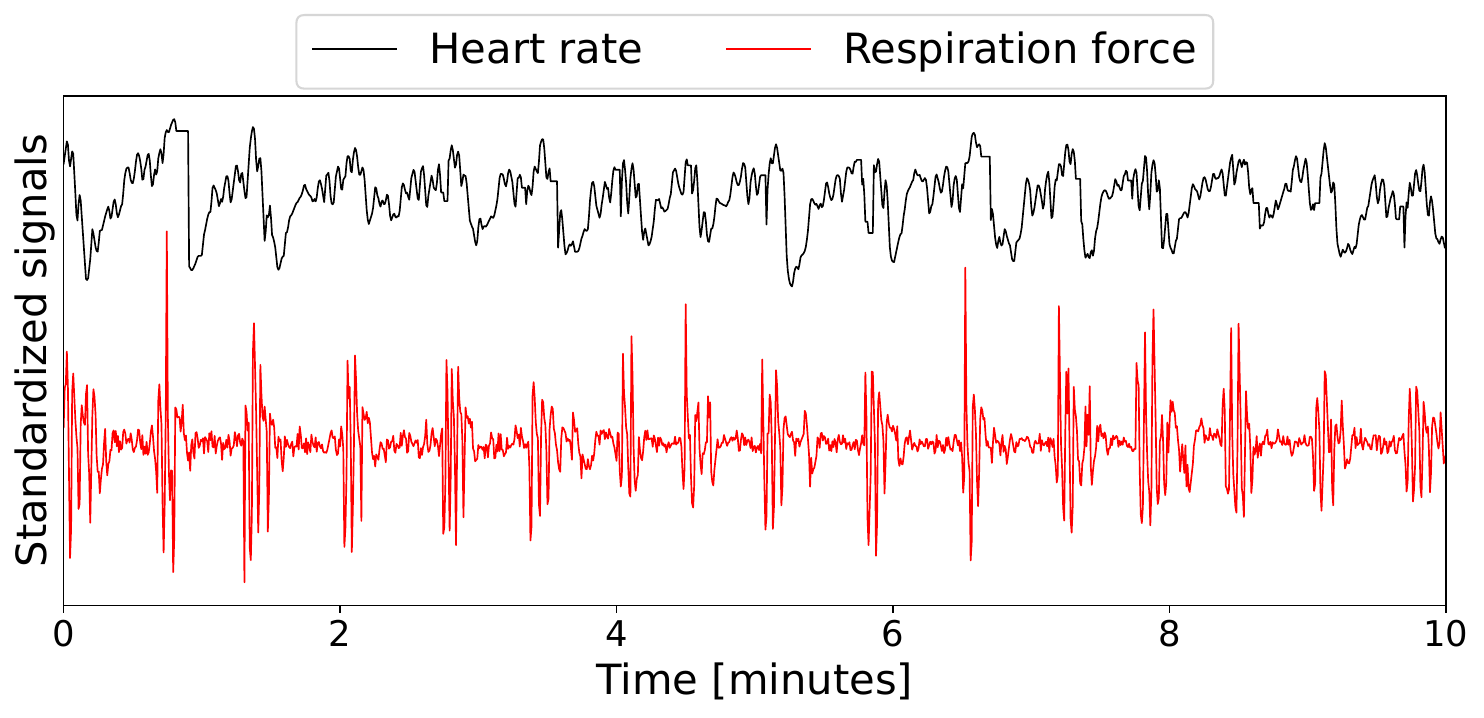}
   \caption{Sampled heartbeat and respiration force time series from the Santa Fe dataset shown over 10 minutes.}
   \label{fig:signals_real}
\end{figure}

The Santa Fe Time Series Competition Data Set B is a multivariate physiological dataset recorded from a patient in a sleep laboratory in Boston, Massachusetts \citep{Rigney1993,ichimaru1999development}. 
It comprises synchronized measurements of heart rate, chest (respiration) volume, and blood oxygen concentration, sampled at 2 Hz (every 0.5 seconds); see \Figref{fig:signals_real}.

To be consistent with previous works that analyze this dataset (e.g., \citet{cactaron2018transfer}), we only consider the chunk of the time series from index 2350 to index 3550.

The TE analysis on the Santa Fe dataset, shown in \Figref{fig:results_real}, reveals consistently higher values from respiration force to heart rate than in the reverse direction, with magnitudes roughly two to three times larger across most of the examined lags. A decay in the transfer of information is observed when conditioning on more three seconds of past respiratory activity, while the reverse direction remains comparatively stable across lags. This asymmetry suggests that the identified directional coupling is robust to the specific lag choice rather than an artifact of delay structure, aligning with prior findings in physiological data \citep{schreiber2000measuring,kaiser2002information,luxembourg2025treet,cactaron2018transfer}. When compared against alternative estimators, also included in \Figref{fig:results_real}, TENDE produces more stable and physiologically interpretable estimates. MINE and Npeet exhibit greater variability and deviations from expected trends. TREET recovers the correct directional asymmetry but with substantially larger error bars, while Tigramite yields estimates that are orders of magnitude smaller than those of all other methods. The declining TE values from respiration to heart rate at longer lags further indicate that extended cardiac history reduces the incremental predictive contribution of the breathing signal, although interpretation must remain cautious given the complexities of coupled physiological systems. Finally, a comparison with AGM was not performed, since its available implementation only supports transfer entropy estimation with a single lag, preventing inclusion under the longer conditioning on the past of the signals setting considered here.
\section{Conclusions}
\label{sec:conclusions}
Quantifying directed information flow in time series remains a central problem in many applications, e.g., in neuroscience, finance, and complex systems analysis. In this paper, we introduced TENDE (Transfer Entropy Neural Diffusion Estimation), a novel approach that leverages score-based diffusion models for flexible and scalable estimation of transfer entropy via conditional mutual information with minimal assumptions.
Experiments on synthetic benchmarks and real-world datasets show that TENDE achieves high accuracy and robustness, outperforming existing neural estimators and other competitors from the state-of-the-art. 
Looking ahead, we aim to extend TENDE to handle nonstationary dynamics and explore amortization across lags to improve efficiency in long time series. While TENDE inherits the computational cost of training diffusion models, it offers a principled and effective framework for transfer entropy estimation, paving the way for more reliable analysis of dependencies in complex dynamical systems. 

\bibliography{refs}
\bibliographystyle{abbrvnat_nourl}


\newpage

\appendix


%
%





%

%

\onecolumn
\aistatstitle{Appendix}

\section{Detailed derivations}
\label{sec:derivs}
\subsection{Entropy by using an auxiliary Gaussian random variable and its estimation}
\label{sec:entr_gauss}
We will first focus on the derivation of \Eqref{eq:ent_kld} and how to use it as the means to estimate the entropy of a random variable. 

Recall that $X$ denotes a $N_\mathrm{x}$-dimensional random variable with density $p_X$, and that  $\varphi_\sigma$ denotes the density of a $N_\mathrm{x}$-dimensional centered Gaussian random variable with covariance $\sigma^2 \mathbf{I}_{N_\mathrm{x}}$. Thus, the KL Divergence between $p_X$ and $\varphi_\sigma$ is given by:
\begin{equation*}
    \begin{aligned}
        \KL{p_X}{\varphi_\sigma} &= \E{x}{p_X} \left[ \log \left( \frac{p_X(x)}{\varphi_\sigma(x)}\right)\right] \\
        &= \E{x}{p_X}\left[\log \left( p_X(x)\right) \right] - \E{x}{p_X}\left[\log \left( \varphi_\sigma(x)\right) \right] \\
        &= \E{x}{p_X}\left[\log \left( p_X(x)\right) \right] - \E{x}{p_X}\left[- \frac{\Vert x \rVert^2}{2 \sigma^2} - \frac{N_\mathrm{x}}{2}  \log \left(2 \pi \sigma^2 \right)\right] \\
        &= \E{x}{p_X}\left[\log \left( p_X(x)\right) \right] - \left(\E{x}{p_X}\left[- \frac{\Vert x \rVert^2}{2 \sigma^2}\right] - \frac{N_\mathrm{x}}{2}  \log \left(2 \pi \sigma^2 \right) \right).\\
    \end{aligned}
\end{equation*}
Thus, rearranging the terms and noticing that $ H(X) = -\E{x}{p_X}\left[\log \left( p_X(x)\right) \right]$ we obtain the desired equality, that is:
\begin{equation*}
    H(X) = \frac{N_\mathrm{x}}{2}  \log \left(2 \pi \sigma^2 \right) + \E{x}{p_X}\left[ \frac{\Vert x \rVert^2}{2 \sigma^2}\right]  - \KL{p_X}{\varphi_\sigma}.
\end{equation*}
With this in mind, we can now use the estimator of the KL Divergence stated in \Eqref{eq:kld_estimator} to estimate the entropy. Notice that there are two unknown densities involved in \Eqref{eq:kld_estimator}, therefore two parametric scores are required. However, that is not the case here since $p_X$ is the only unknown, hence, only a single score network is required to estimate the KL Divergence between $p_X$ and $\varphi_\sigma$. It is important to keep in mind that if we construct the following diffusion process,
\begin{equation*}
    \begin{cases}
        \begin{aligned}
            dX_t &= f_t X_t dt + g_t dW_t \\
            X_0 &\sim \varphi_\sigma,
        \end{aligned}
    \end{cases}
\end{equation*}
the score function associated with $X_t$ is known and is given by $S^{\varphi_{\sigma_{X_t}}}(x) = -\frac{x}{\chi_t}$, where $\chi_t = \left( k_t^2 \sigma^2 + k_t^2 \int_0^t \frac{g_s^2}{k_s^2} ds \right)$ with $k_t = \exp\left \{\int_0^t f_s ds \right \}$. Replacing $q$ by $\varphi_\sigma$ yields:
\begin{equation*}
    \begin{aligned}
        \KL{p_X}{\varphi_\sigma} &= \int_0^{T} \frac{g_t^2}{2}   \E{x}{p_{X_t}} \left[\left \lVert \Score{p}{X_t}(x) - S^{\varphi_{\sigma_{X_t}}}(x) \right \rVert^2 \right] dt + D_{KL} \left[p_{X_T} \vert \vert \varphi_{\sigma_{X_T}} \right] \\
        &=\int_0^{T} \frac{g_t^2}{2}   \E{x}{p_{X_t}} \left[\left \lVert \Score{p}{X_t}(x) - S^{\varphi_{\sigma_{X_t}}}(x) \right \rVert^2 \right] dt + \KL{\varphi_1}{\varphi_{\sqrt{\chi_T}}} \\
        &=\int_0^{T} \frac{g_t^2}{2}   \E{x}{p_{X_t}} \left[\left \lVert \Score{p}{X_t}(x) - S^{\varphi_{\sigma_{X_t}}}(x) \right \rVert^2 \right] dt + \frac{N_\mathrm{x}}{2} \left( \log(\chi_T) - 1 + \frac{1}{\chi_T}\right)\\
        &\simeq e(p_X, \varphi_\sigma) + \frac{N_\mathrm{x}}{2}\left( \log(\chi_T) - 1 + \frac{1}{\chi_T}\right).
    \end{aligned}
\end{equation*}
The first equality is simply \Eqref{eq:kld_diff}; the second equality follows from the fact that using the variance preserving stochastic differential equation, $p_{X_T} \simeq \varphi_1$ for $T$ large enough. Similarly, we have that when $X_0$ is sampled from $\varphi_\sigma$ the random variable $X_T \sim \varphi_{\sqrt{\chi_T}}$, thus $D_{KL} \left[p_{X_T} \vert \vert \varphi_{\sigma_{X_T}} \right] = \KL{\varphi_1}{\varphi_{\sqrt{\chi_T}}}$. The third equality arises due to the fact that $\KL{\varphi_1}{\varphi_{\sqrt{\chi_T}}}$ is available in closed form, and the last equality is simply obtained by replacing the first term with its respective approximation. Finally, we have:
\begin{equation*}
    \begin{aligned}
        H(X) &= \frac{N_\mathrm{x}}{2}  \log \left(2 \pi \sigma^2 \right) + \E{x}{p_X}\left[ \frac{\Vert x \rVert^2}{2 \sigma^2}\right] - \KL{p_X}{\varphi_\sigma} \\
        &\simeq \frac{N_\mathrm{x}}{2}  \log \left(2 \pi \sigma^2 \right) + \E{x}{p_X}\left[ \frac{\Vert x \rVert^2}{2 \sigma^2}\right] - \left[ e(p_X, \varphi_\sigma) + \frac{N_\mathrm{x}}{2}\left( \log(\chi_T) - 1 + \frac{1}{\chi_T}\right)\right] \\
        &= \frac{N_\mathrm{x}}{2}  \log \left(2 \pi \sigma^2 \right) + \E{x}{p_X}\left[ \frac{\Vert x \rVert^2}{2 \sigma^2}\right] - e(p_X, \varphi_\sigma) - \frac{N_\mathrm{x}}{2}\left( \log(\chi_T) - 1 + \frac{1}{\chi_T}\right). \\
        &= H(X; \sigma) 
    \end{aligned}
\end{equation*}
\subsection{Derivation of TE estimators}
\label{sec:derivs_ests}
\subsubsection{TE as expected KL Divergence}
Deriving the estimator proposed in \Eqref{eq:cmihat1} is a straightforward application of \Eqref{eq:te_kld} and the fact that $e(\cdot, \cdot)$ is our estimator for KL Divergence (see \Secref{sec:overview_scores}), thus we have:
\begin{equation*}
    \begin{aligned}
        I(X, Y \vert Z) &= \  \E{[y, z]}{p_{Y, Z}} \left[ \KL{p_{X_{y, z}}}{p_{X_z}} \right] \\
        &\simeq \E{[y, z]}{p_{Y, Z}} \left[ e(p_{X_{y, z}}, p_{X_z}) \right].
    \end{aligned}
\end{equation*}
\subsubsection{TE as difference of conditional entropies}
Recall that $I(X; Y \vert Z) = \mathcal{H}(X \vert Z) - \mathcal{H} (X \vert Y, Z)$. Using \Eqref{eq:ent_kld_hat} we have:
\begin{equation*}
        \begin{aligned}
            \mathcal{H} (X \vert Y, Z) &\simeq  \E{[y, z]}{p_{Y, Z}} \left[ \frac{N_\mathrm{x}}{2} \log(2 \pi \sigma^2) + \E{x}{p_{X_{y, z}}} \left[ \frac{\lVert x \rVert^2}{2 \sigma^2}\right] - e(p_{X_{y, z}}, \varphi_\sigma) - \frac{N_\mathrm{x}}{2} \left( \log(\chi_T) - 1 + \frac{1}{\chi_T} \right) \right] \\
            &= \frac{N_\mathrm{x}}{2} \log(2 \pi \sigma^2) + \E{x}{p_X} \left[ \frac{\lVert x \rVert^2}{2 \sigma^2}\right] - \E{[y, z]}{p_{Y, Z}} \left[ e(p_{X_{y, z}}, \varphi_\sigma) \right] - \frac{N_\mathrm{x}}{2} \left( \log(\chi_T) - 1 + \frac{1}{\chi_T} \right).
        \end{aligned}
\end{equation*}
In a similar fashion, it is possible to obtain the following:
\begin{equation*}
    \mathcal{H} (X \vert Z) \simeq \frac{N_\mathrm{x}}{2} \log(2 \pi \sigma^2) + \E{x}{p_X} \left[ \frac{\lVert x \rVert^2}{2 \sigma^2}\right] - \E{z}{p_Z} \left[ e(p_{X_{z}}, \varphi_\sigma) \right] - \frac{N_\mathrm{x}}{2} \left( \log(\chi_T) - 1 + \frac{1}{\chi_T} \right).
\end{equation*}
Thus, it immediately follows that: 
\begin{equation*}
    \begin{split}
    I(X; Y \vert Z) &= \mathcal{H}(X \vert Z) - \mathcal{H} (X \vert Y, Z) \\
    &\simeq \E{[y, z]}{p_{Y, Z}} \left[ e(p_{X_{y, z}}, \varphi_\sigma) \right] - \E{z}{p_Z} \left[ e(p_{X_{z}}, \varphi_\sigma) \right] \text{.}
    \end{split}
\end{equation*}
\subsubsection{TE as difference of mutual informations}
We leverage the representation of conditional mutual information as the difference of mutual informations in the case of the estimator proposed in \Eqref{eq:mis}, that is $ I(X; Y \vert Z) = I(X; [Y, Z]) - I(X; Z)$. Furthermore we represent the mutual informations as the expectation over KL Divergencies as follows:
\begin{equation*}
    \begin{aligned}
        I(X; Y \vert Z) &= I(X; [Y, Z]) - I(X; Z) \\
        &= \E{[y, z]}{p_{Y, Z}} \left[ \KL{p_{X_{y, z}}}{p_{X}} \right] - \E{z}{p_{Z}} \left[ \KL{p_{X_{z}}}{p_{X}} \right] \\
        &\simeq \E{[y, z]}{p_{Y, Z}} \left[ e(p_{X_{y, z}}, {p_{X}}) \right] - \E{z}{p_{Z}} \left[ e(p_{X_{z}}, p_{X}) \right] \text{.}
    \end{aligned}
\end{equation*}

\subsection{Approximation error}
\label{sec:approx_error}
We now discuss the quality of the approximation $e(p, q) \approx D_{KL}[p \| q]$ introduced in \Eqref{eq:kld_estimator}. Recall from \Eqref{eq:kld_diff} that the exact KL divergence decomposes as
\begin{equation*}
    D_{KL}[p \| q] = \underbrace{\int_0^{T} \frac{g_t^2}{2} \E{x}{p_{X_t}} \left[ \left\lVert \Score{p}{X_t}(x) - \Score{q}{X_t}(x) \right\rVert^2 \right] dt}_{\text{score difference term}} + \underbrace{D_{KL}[p_{X_T} \| q_{X_T}]}_{\text{terminal divergence}}.
\end{equation*}
Since $e(p, q)$ replaces the true scores with their parametric approximations in the first term, the estimation error is given by
\begin{equation*}
    e(p, q) - D_{KL}[p \| q] = d - D_{KL}[p_{X_T} \| q_{X_T}],
\end{equation*}
where, defining the score errors $\epsilon^{p}_t(x) := \Score{p}{X_t}(x; \theta^\star) - \Score{p}{X_t}(x)$ and $\epsilon^{q}_t(x) := \Score{q}{X_t}(x; \theta^\star) - \Score{q}{X_t}(x)$, the term $d$ has the form \citep{franzese2023minde}
\begin{equation*}
    d = \int_0^{T} \frac{g_t^2}{2} \E{x}{p_{X_t}} \left[ \left\lVert \epsilon^{p}_t(x) - \epsilon^{q}_t(x) \right\rVert^2 + 2 \left\langle \Score{p}{X_t}(x) - \Score{q}{X_t}(x), \ \epsilon^{p}_t(x) - \epsilon^{q}_t(x) \right\rangle \right] dt.
\end{equation*}
Two observations are worth noting. First, $d$ is neither necessarily positive nor negative, so the estimator $e(p, q)$ is neither an upper nor a lower bound of the true KL divergence. This frees our approach from the pessimistic results of \citet{mcallester2020formal} that affect variational estimators. Second, common-mode score errors cancel: if $\epsilon^{p}_t(x) = \epsilon^{q}_t(x)$, then $d = 0$ regardless of the individual error magnitudes.

Regarding the terminal divergence $D_{KL}[p_{X_T} \| q_{X_T}]$, for the Variance Preserving schedule used in this work, the contraction properties of the diffusion semigroup \citep{collet2008logarithmic} ensure that both $p_{X_T}$ and $q_{X_T}$ converge to the same stationary distribution as $T$ grows, rendering this term numerically negligible for the values of $T$ used in practice.

\newpage

\section{Proofs}
\label{sec:proofs}

\subsection{Invariance of the TE when stacking redundant dimensions}
\label{sec:pf1}
Recall that in \Secref{sec:estimation_benchmark} we defined the redundant setting as the stacking of $d$ redundant dimensions onto both $x_t$ and $y_t$. More generally, we could consider two time series $\tilde{x}_t$ and $\tilde{y}_t$ defined as follows: 
\begin{equation*}
    \begin{cases}
        \begin{aligned}
            \tilde{x}_t &= \left[x_t, \varepsilon^{x}_{t, 1}, \dots \varepsilon^{x}_{t, d_x} \right] \\
            \tilde{y}_t &= \left[y_t,  \varepsilon^{y}_{t, 1}, \dots \varepsilon^{y}_{t, d_y} \right].
        \end{aligned}
    \end{cases}
\end{equation*}

The redundant dimensions $\varepsilon^{x}_{t,i}$ and $\varepsilon^{y}_{t,j}$ are taken to be mutually independent collections. In particular, for all $t, t', i, i', j, j'$ we have $\varepsilon^{x}_{t,i} \perp\!\!\!\perp \varepsilon^{x}_{t',i'}$, $\varepsilon^{y}_{t,j} \perp\!\!\!\perp \varepsilon^{y}_{t',j'}$, and $\varepsilon^{x}_{t,i} \perp\!\!\!\perp \varepsilon^{y}_{t',j'}$. Moreover, each of these redundant components is independent of the original processes, i.e., $\{\varepsilon^{x}_{t,i}\}_{t,i} \perp\!\!\!\perp (x_t, y_t)$ and $\{\varepsilon^{y}_{t,j}\}_{t,j} \perp\!\!\!\perp (x_t, y_t)$. To avoid clutter, we drop the subscripts on the distribution functions as well as the distribution functions in the expectations. That being said, let $k, \ell \in \mathbb{N}$ be the lags and construct $\mathbf{\tilde{x}}_{t -k}$ and $\mathbf{\tilde{y}}_{t - \ell}$ as defined in \Secref{sec:te_def}. Also, define $\varepsilon^{x}_{t} = \left[\varepsilon^{x}_{t, 1}, \dots \varepsilon^{x}_{t, d_x} \right]$, and define $\varepsilon^{y}_{t}$ similarly. First, consider the distribution of $\tilde{y}_t$ and $\mathbf{\tilde{x}}_{t - k}$ conditioned on $\mathbf{\tilde{y}}_{t - \ell}$. 
We can see that:
\begin{equation*}
    \begin{aligned}
        p\left(\tilde{y}_t, \mathbf{\tilde{x}}_{t - k} \vert \mathbf{\tilde{y}}_{t - \ell} \right) &= \frac{p\left(\tilde{y}_t, \mathbf{\tilde{x}}_{t - k}, \mathbf{\tilde{y}}_{t - \ell} \right)}{p\left(\mathbf{\tilde{y}}_{t - \ell} \right)} \\
        &= \frac{p \left( y_t, \mathbf{x}_{t - k}, \mathbf{y}_{t - \ell}, \varepsilon^{y}_{t}, \varepsilon^{x}_{t - k}, \varepsilon^{y}_{t - \ell} \right)}{p \left( \mathbf{y}_{t - \ell}, \varepsilon^{y}_{t - \ell} \right)} \\ 
        &=\frac{p \left( y_t, \mathbf{x}_{t - k}, \mathbf{y}_{t - \ell} \right) p\left(\varepsilon^{y}_{t}\right) p\left(\varepsilon^{x}_{t - k}\right) p \left(\varepsilon^{y}_{t - \ell} \right)}{p \left( \mathbf{y}_{t - \ell} \right) p \left(\varepsilon^{y}_{t - \ell} \right)} \\
        &= p \left( y_t, \mathbf{x}_{t - k} \vert \mathbf{y}_{t - \ell} \right) p\left(\varepsilon^{y}_{t}\right) p\left(\varepsilon^{x}_{t - k}\right),
    \end{aligned}
\end{equation*}
where the third equality comes from the construction of the system $\left[\tilde{x}_t, \tilde{y}_t\right]$ and the other equalities are immediate to deduce. Using similar arguments, it is possible to show that $p\left( \mathbf{\tilde{x}}_{t - k} \vert \mathbf{\tilde{y}}_{t - \ell} \right) = p\left( \mathbf{x}_{t - k} \vert \mathbf{y}_{t - \ell} \right) p \left( \varepsilon^{x}_{t - k}\right)$ and $p\left( \tilde{y}_{t} \vert \mathbf{\tilde{y}}_{t - \ell} \right) = p\left( y_{t} \vert \mathbf{y}_{t - \ell} \right) p \left( \varepsilon^{y}_{t}\right)$, thus
\begin{equation}
\label{eq:no_redundant}
    \frac{p\left(\tilde{y}_t, \mathbf{\tilde{x}}_{t - k} \vert \mathbf{\tilde{y}}_{t - \ell} \right)}{p\left(\mathbf{\tilde{x}}_{t - k} \vert \mathbf{\tilde{y}}_{t - \ell} \right) p\left(\tilde{y}_t \vert \mathbf{\tilde{y}}_{t - \ell} \right)} = \frac{p \left( y_t, \mathbf{x}_{t - k} \vert \mathbf{y}_{t - \ell} \right)}{p \left(\mathbf{x}_{t - k} \vert \mathbf{y}_{t - \ell} \right) p \left( y_t \vert \mathbf{y}_{t - \ell} \right)}.
\end{equation}
Finally, consider the transfer entropy from $\tilde{x}$ to $\tilde{y}$
\begin{equation*}
    \begin{aligned}
        \mathrm{TE}_{\tilde{X} \to \tilde{Y}}(k,\ell) &= \mathbb{E}_{\mathbf{\tilde{y}}_{t - k}} \left[ \KL{p\left(\tilde{y}_t, \mathbf{\tilde{x}}_{t - k} \vert \mathbf{\tilde{y}}_{t - \ell} \right)}{p\left(\mathbf{\tilde{x}}_{t - k} \vert \mathbf{\tilde{y}}_{t - \ell} \right) p\left(\tilde{y}_t \vert \mathbf{\tilde{y}}_{t - \ell} \right)}\right] \\
        &=\mathbb{E}_{\tilde{y}_t, \mathbf{\tilde{x}}_{t - k}, \mathbf{\tilde{y}}_{t - \ell}} \left[ \log \left( \frac{p\left(\tilde{y}_t, \mathbf{\tilde{x}}_{t - k} \vert \mathbf{\tilde{y}}_{t - \ell} \right)}{p\left(\mathbf{\tilde{x}}_{t - k} \vert \mathbf{\tilde{y}}_{t - \ell} \right) p\left(\tilde{y}_t \vert \mathbf{\tilde{y}}_{t - \ell} \right)}\right) \right] \\
        &= \mathbb{E}_{\tilde{y}_t, \mathbf{\tilde{x}}_{t - k}, \mathbf{\tilde{y}}_{t - \ell}} \left[ \log \left( \frac{p \left( y_t, \mathbf{x}_{t - k} \vert \mathbf{y}_{t - \ell} \right)}{p \left(\mathbf{x}_{t - k} \vert \mathbf{y}_{t - \ell} \right) p \left( y_t \vert \mathbf{y}_{t - \ell} \right)}\right)\right] \\
        &= \mathbb{E}_{{y}_t, \mathbf{{x}}_{t - k}, \mathbf{{y}}_{t - \ell}} \left[ \log \left( \frac{p \left( y_t, \mathbf{x}_{t - k} \vert \mathbf{y}_{t - \ell} \right)}{p \left(\mathbf{x}_{t - k} \vert \mathbf{y}_{t - \ell} \right) p \left( y_t \vert \mathbf{y}_{t - \ell} \right)}\right)\right] \\
        &= \mathrm{TE}_{X \to Y}(k,\ell).
    \end{aligned}
\end{equation*}
Where the first two equalities follow from the definition of TE, the third equality is consequence of \Eqref{eq:no_redundant}, furthermore, the forth equality follows from the fact that the expression at hand does not depend on the redundant dimensions anymore. The last equality follows from the definition of TE.

The proof in the other direction is identical.
\newpage

\subsection{Additivity of the TE when independent components are stacked}
\label{sec:pf2}
in \Secref{sec:estimation_benchmark} we defined the stacking setting as stacking of $d$ independent replicates of the processes $x_t$ and $y_t$ in such a way that dependence exists only between corresponding components. More generally consider $\{ x_{t, i} \}$ and $\{ y_{t, i}\}$. The components $x_{t,i}$ and $x_{t',j}$ are assumed to be independent for all $t, t', i, j$, and analogously $y_{t,i}$ and $y_{t',j}$ are independent for all indices. The only dependence between the two processes arises when the second sub-index coincides, that is, $x_{t,i}$ and $y_{t',i}$ may be dependent, while $x_{t,i}$ and $y_{t',j}$ are independent for $i \neq j$. With these assumptions, we construct the series $\tilde{x}_t$ and $\tilde{y}_t$ as:
\begin{equation*}
    \begin{cases}
        \begin{aligned}
            \tilde{x}_t &= \left[x_{t, 1}, \dots x_{t, d} \right] \\
            \tilde{y}_t &= \left[y_{t, 1}, \dots y_{t, d} \right],
        \end{aligned}
    \end{cases}
\end{equation*}
As in \Secref{sec:pf1}, we avoid cluttering the notation by dropping the subscripts on the distribution functions and the distribution functions in the expectations. Similarly, let $k, \ell \in \mathbb{N}$ be the lags and construct $\mathbf{\tilde{x}}_{t -k}$ and $\mathbf{\tilde{y}}_{t - \ell}$ as defined in \Secref{sec:te_def}. First, consider the distribution of $\tilde{y}_t$ and $\mathbf{\tilde{x}}_{t - k}$ conditioned on $\mathbf{\tilde{y}}_{t - \ell}$, thus we can see that
\begin{equation*}
    \begin{aligned}
        p\left(\tilde{y}_t, \mathbf{\tilde{x}}_{t - k} \vert \mathbf{\tilde{y}}_{t - \ell} \right) &= \frac{p\left(\tilde{y}_t, \mathbf{\tilde{x}}_{t - k}, \mathbf{\tilde{y}}_{t - \ell} \right)}{p\left(\mathbf{\tilde{y}}_{t - \ell} \right)} \\
        &= \frac{p \left( y_{t, 1}, \mathbf{x}_{t - k, 1}, \mathbf{y}_{t - \ell, 1}, \dots, y_{t, d}, \mathbf{x}_{t - k, d}, \mathbf{y}_{t - \ell, d} \right)}{p \left( \mathbf{y}_{t - \ell, 1}, \dots,  \mathbf{y}_{t - \ell, 1} \right)} \\ 
        &= \frac{\prod_{j = 1}^{d} p \left( y_{t, j}, \mathbf{x}_{t - k, j}, \mathbf{y}_{t - \ell, j}\right)}{\prod_{j = 1}^{d} p \left( \mathbf{y}_{t - \ell, j}\right)} \\ 
        &= \prod_{j = 1}^{d} p \left( y_{t, j}, \mathbf{x}_{t - k, j} \vert \mathbf{y}_{t - \ell, j}\right).
    \end{aligned}
\end{equation*}
The first and second equalities are immediate and the third one arises from the design of the system; the forth equality is immediate as well. Using the same arguments, it is possible to obtain similar decompositions for the other quantities of interest, namely, $p\left( \mathbf{\tilde{x}}_{t - k} \vert \mathbf{\tilde{y}}_{t - \ell} \right)$ and $p\left( \tilde{y}_{t} \vert \mathbf{\tilde{y}}_{t - \ell} \right)$. That is, $p\left( \mathbf{\tilde{x}}_{t - k} \vert \mathbf{\tilde{y}}_{t - \ell} \right) = \prod_{j = 1}^{d}p\left( \mathbf{x}_{t - k, j} \vert \mathbf{y}_{t - \ell, j} \right)$ and $p\left( \tilde{y}_{t} \vert \mathbf{\tilde{y}}_{t - \ell} \right) = \prod_{j = 1}^{d} p\left( y_{t, j} \vert \mathbf{y}_{t - \ell, j} \right)$, hence
\begin{equation}
\label{eq:no_stacking}
    \frac{p\left(\tilde{y}_t, \mathbf{\tilde{x}}_{t - k} \vert \mathbf{\tilde{y}}_{t - \ell} \right)}{p\left(\mathbf{\tilde{x}}_{t - k} \vert \mathbf{\tilde{y}}_{t - \ell} \right) p\left(\tilde{y}_t \vert \mathbf{\tilde{y}}_{t - \ell} \right)} = \prod_{j = 1}^{d} \frac{p \left( y_{t, j}, \mathbf{x}_{t - k, j} \vert \mathbf{y}_{t - \ell, j}\right)}{p \left(\mathbf{x}_{t - k, j} \vert \mathbf{y}_{t - \ell, j} \right) p \left( y_{t, j} \vert \mathbf{y}_{t - \ell, j} \right)}.
\end{equation}
Finally, consider the transfer entropy from $\tilde{x}$ to $\tilde{y}$
\begin{equation*}
    \begin{aligned}
        \mathrm{TE}_{\tilde{X} \to \tilde{Y}}(k,\ell) &= \mathbb{E}_{\mathbf{\tilde{y}}_{t - k}} \left[ \KL{p\left(\tilde{y}_t, \mathbf{\tilde{x}}_{t - k} \vert \mathbf{\tilde{y}}_{t - \ell} \right)}{p\left(\mathbf{\tilde{x}}_{t - k} \vert \mathbf{\tilde{y}}_{t - \ell} \right) p\left(\tilde{y}_t \vert \mathbf{\tilde{y}}_{t - \ell} \right)}\right] \\
        &=\mathbb{E}_{\tilde{y}_t, \mathbf{\tilde{x}}_{t - k}, \mathbf{\tilde{y}}_{t - \ell}} \left[ \log \left( \frac{p\left(\tilde{y}_t, \mathbf{\tilde{x}}_{t - k} \vert \mathbf{\tilde{y}}_{t - \ell} \right)}{p\left(\mathbf{\tilde{x}}_{t - k} \vert \mathbf{\tilde{y}}_{t - \ell} \right) p\left(\tilde{y}_t \vert \mathbf{\tilde{y}}_{t - \ell} \right)}\right) \right] \\
        &= \mathbb{E}_{\tilde{y}_t, \mathbf{\tilde{x}}_{t - k}, \mathbf{\tilde{y}}_{t - \ell}} \left[ \log \left( \prod_{j = 1}^{d} \frac{p \left( y_{t, j}, \mathbf{x}_{t - k, j} \vert \mathbf{y}_{t - \ell, j}\right)}{p \left(\mathbf{x}_{t - k, j} \vert \mathbf{y}_{t - \ell, j} \right) p \left( y_{t, j} \vert \mathbf{y}_{t - \ell, j} \right)} \right)\right] \\
        &= \sum_{j = 1}^{d} \mathbb{E}_{\tilde{y}_t, \mathbf{\tilde{x}}_{t - k}, \mathbf{\tilde{y}}_{t - \ell}} \left[ \log \left( \frac{p \left( y_{t, j}, \mathbf{x}_{t - k, j} \vert \mathbf{y}_{t - \ell, j}\right)}{p \left(\mathbf{x}_{t - k, j} \vert \mathbf{y}_{t - \ell, j} \right) p \left( y_{t, j} \vert \mathbf{y}_{t - \ell, j} \right)} \right)\right] \\
        &= \sum_{j = 1}^{d} \mathbb{E}_{{y}_{t, j}, \mathbf{{x}}_{t - k, j}, \mathbf{{y}}_{t - \ell, j}} \left[ \log \left( \frac{p \left( y_{t, j}, \mathbf{x}_{t - k, j} \vert \mathbf{y}_{t - \ell, j}\right)}{p \left(\mathbf{x}_{t - k, j} \vert \mathbf{y}_{t - \ell, j} \right) p \left( y_{t, j} \vert \mathbf{y}_{t - \ell, j} \right)} \right)\right] \\
        &= \sum_{j = 1}^{d} \mathrm{TE}_{X_j \to Y_j}(k,\ell).
    \end{aligned}
\end{equation*}
Here the first two equalities follow from the definition of TE, and the third equality is consequence of \Eqref{eq:no_stacking}. 
The forth equality is immediate, and the fifth equality follows from the fact that the expression inside the sum only depends on the $j-$th process. Finally, the last equality follows from the definition of TE.

The proof in the other direction is identical.
\newpage

\section{Further details on the synthetic benchmark and additional experiments}
\label{sec:addexps}
\subsection{Details on the experimental benchmark}
All the stochastic systems analyzed in this study were simulated using the publicly available code at the following link\footnote{\href{https://github.com/dkornai/TE_datasim}{TE\_datasim}} provided by \citet{kornai2025agm}, ensuring consistency with the original experimental setup. The implementations of the NPEET\footnote{\href{https://github.com/gregversteeg/NPEET?tab=readme-ov-file}{NPEET}} \citep{steeg_information-theoretic_2013}, AGM\footnote{\href{https://github.com/dkornai/AGM-TE/tree/main/agm_te}{AGM\_TE}} \citep{kornai2025agm},  TREET\footnote{\href{https://github.com/omerlux/TREET}{TREET}} \citep{luxembourg2025treet}, and Tigramite \citep{runge2019detecting}\footnote{\href{https://github.com/jakobrunge/tigramite}{Tigramite}} were used with their default settings.
Furthermore, the MINE-based transfer entropy estimator was implemented by leveraging the formulation of transfer entropy as the difference between two mutual information terms (see \Eqref{eq:mis}), which allows for the application of neural estimation techniques originally developed for mutual information. In this case, the implementation was obtained using the Benchmarking Mutual Information package\footnote{\href{https://pypi.org/project/benchmark-mi/}{Benchmarking Mutual Information}}. The implementation of TENDE was based on the publicly available code for MINDE\footnote{\href{https://github.com/MustaphaBounoua/minde}{MINDE}}, adapting it to the transfer entropy estimation framework. For the TENDE variants that include $\sigma$ as a hyperparameter, we set $\sigma = 1$, following the configuration adopted in \citet{franzese2023minde}, where this value was shown to yield stable and reliable performance across a variety of stochastic systems. Furthermore, as in \citet{franzese2023minde}, importance sampling was employed during the estimation of transfer entropy. Finally, for all models, the default hyperparameters provided in their original implementations were used during training to ensure fair and reproducible comparisons. Tigramite was excluded from the stacking benchmarks beyond 10 dimensions due to scalability constraints. For the 70-dimensional benchmarks with $T = 50000$, the number of training epochs for TREET was reduced due to numerical instabilities (NaN losses) encountered under the default configuration.

\vfill

\subsection{Beyond Gaussian benchmarks}
In this section, we evaluate TENDE and the competitors we considered in \Secref{sec:experiment:synthetic} across more challenging distributions. MI-invariant transformations are applied to the data to construct such settings. Since TE can be written in terms of MI, the invariance of MI implies invariance of TE, that is, applying MI-invariant transformations to the data leaves the ground truth value of the TE unchanged.
\vfill
\newpage

\subsubsection{Half cube}
Inspired by the work of \citet{franzese2023minde} and \citet{bounoua2024s}, we consider the MI-invariant transformation defined as $x \mapsto\; x \sqrt{|x|}$.

\begin{figure}[h!]
    \centering
    \includegraphics[width=0.48\textwidth]{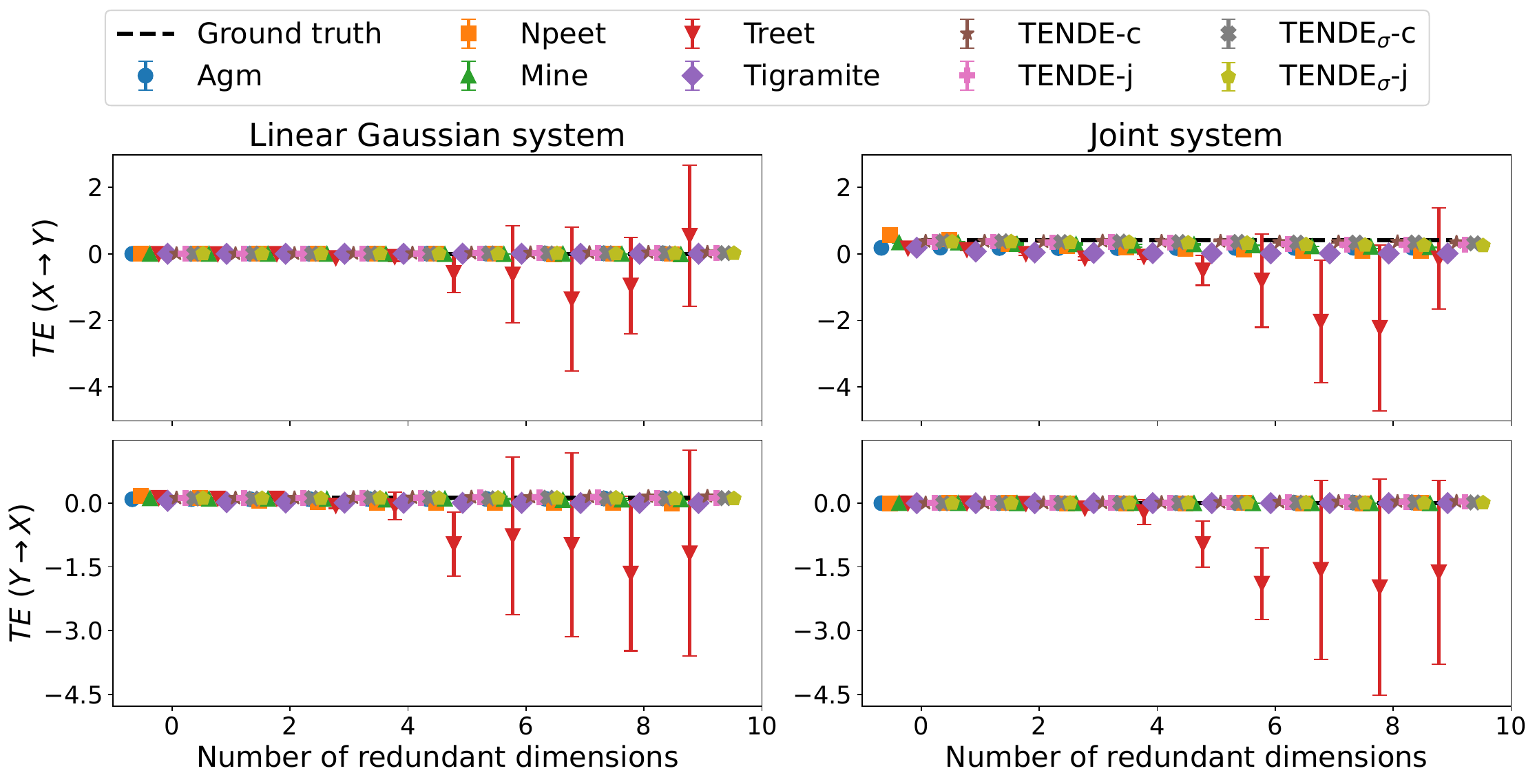}
    \includegraphics[width=0.48\textwidth]{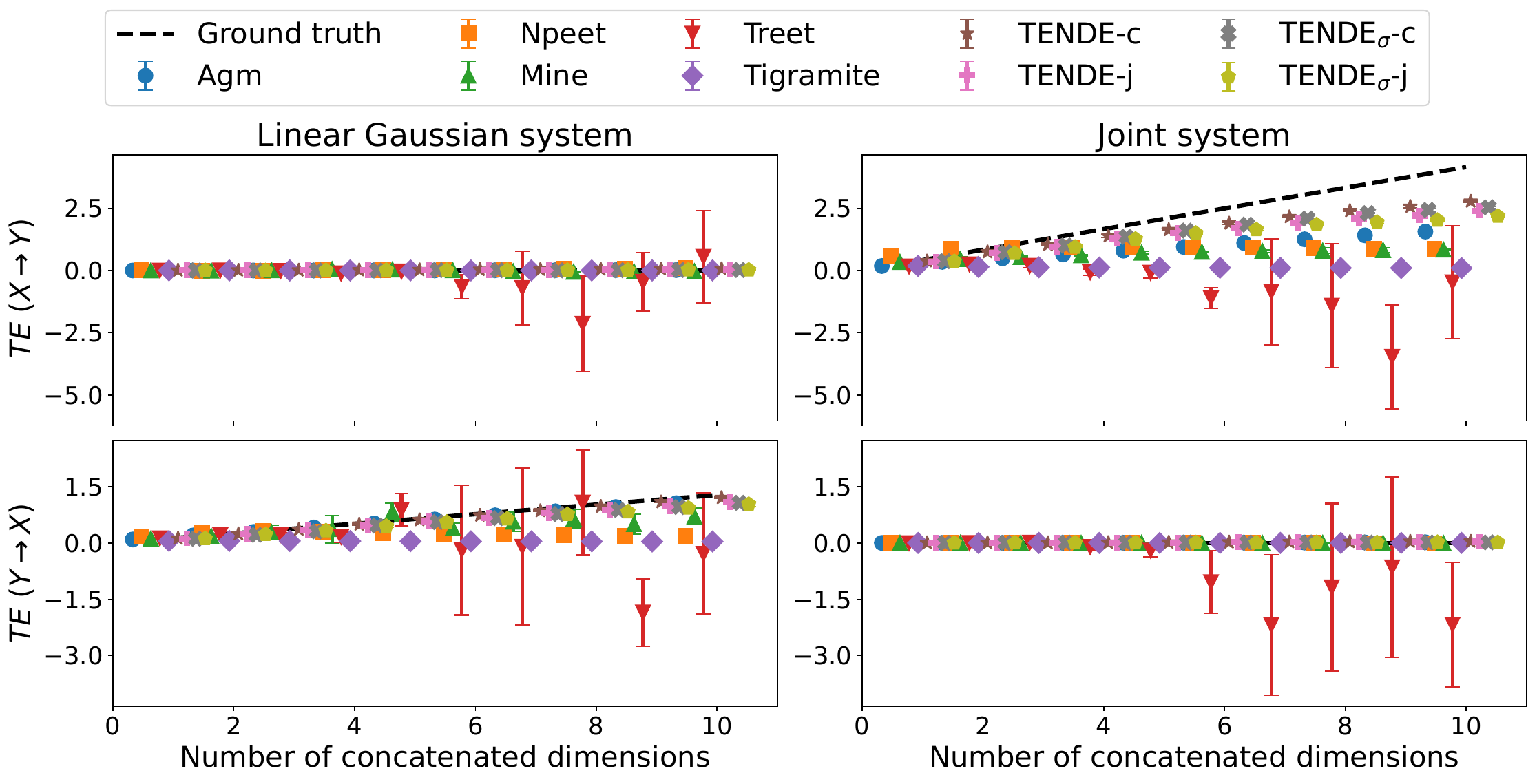}
    \includegraphics[width=0.48\textwidth]{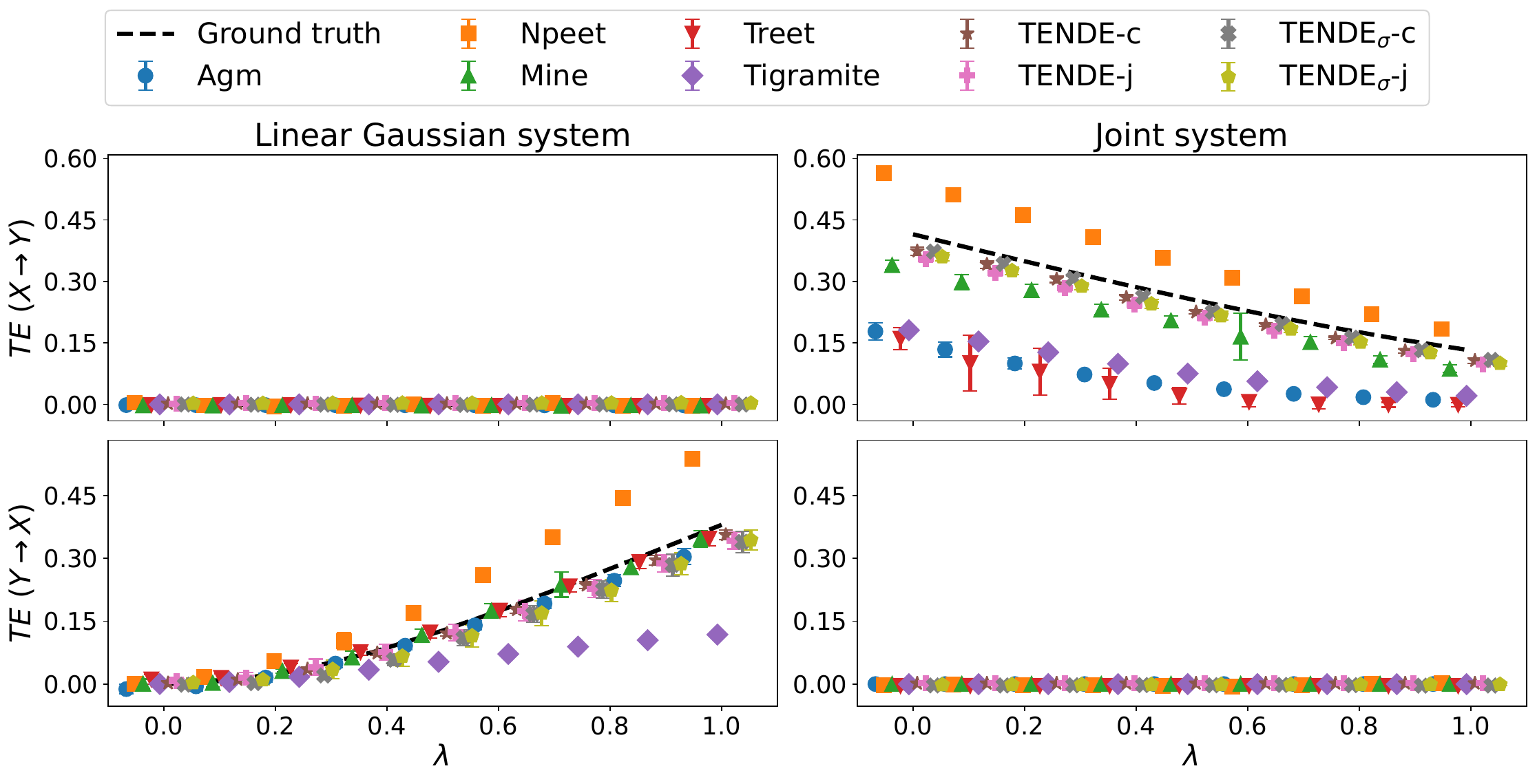}
        \caption{Estimated transfer entropy for the linear Gaussian and joint systems under redundant and linear stacking (top) and varying coupling strength $\lambda$ (bottom). Both systems are modified using the half-cube mapping.}
    \label{fig:half_cube}
\end{figure}
Across all configurations, the TENDE estimators continue to align closely with the analytical ground truth and exhibit consistent behavior across different regimes. In the redundant stacking setting (top-left), where independent noise dimensions are added, TENDE maintains stable estimates across varying numbers of redundant dimensions. In contrast, Treet produces negative estimates with large variance at higher dimensions, and Tigramite consistently underestimates the transfer entropy. In the linear stacking scenario (top-right), TENDE accurately captures the expected linear trend, whereas alternative estimators tend to underestimate the magnitude of transfer entropy and show noticeable bias as dimensionality grows; Treet again exhibits high instability. For the simple coupling system (bottom), TENDE maintains close agreement with the ground truth, while Npeet and Tigramite deviate at higher coupling values, and Treet shows substantial variability across the range of $\lambda$.
\newpage
\subsubsection{CDF}
Following again \citet{franzese2023minde} and \citet{bounoua2024s}, the second MI-invariant transformation we consider is $x \mapsto \Phi^{-1} (x)$, where $\Phi(\cdot)$ denotes the cumulative distribution function (CDF) of a standard Gaussian random variable, mapping all the data to the interval $[0, 1]$.

\begin{figure}[h!]
    \centering
    \includegraphics[width=0.48\textwidth]{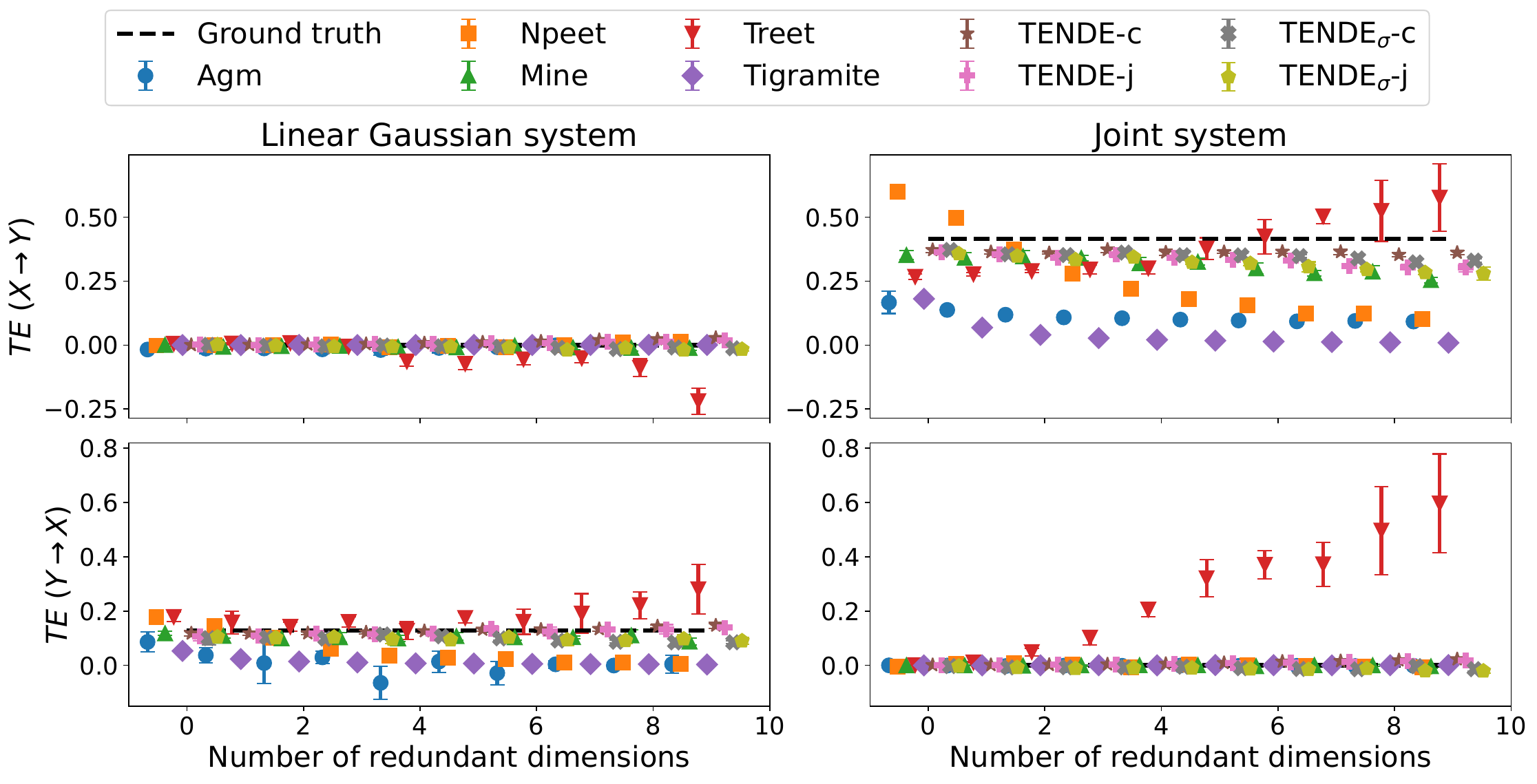}
    \includegraphics[width=0.48\textwidth]{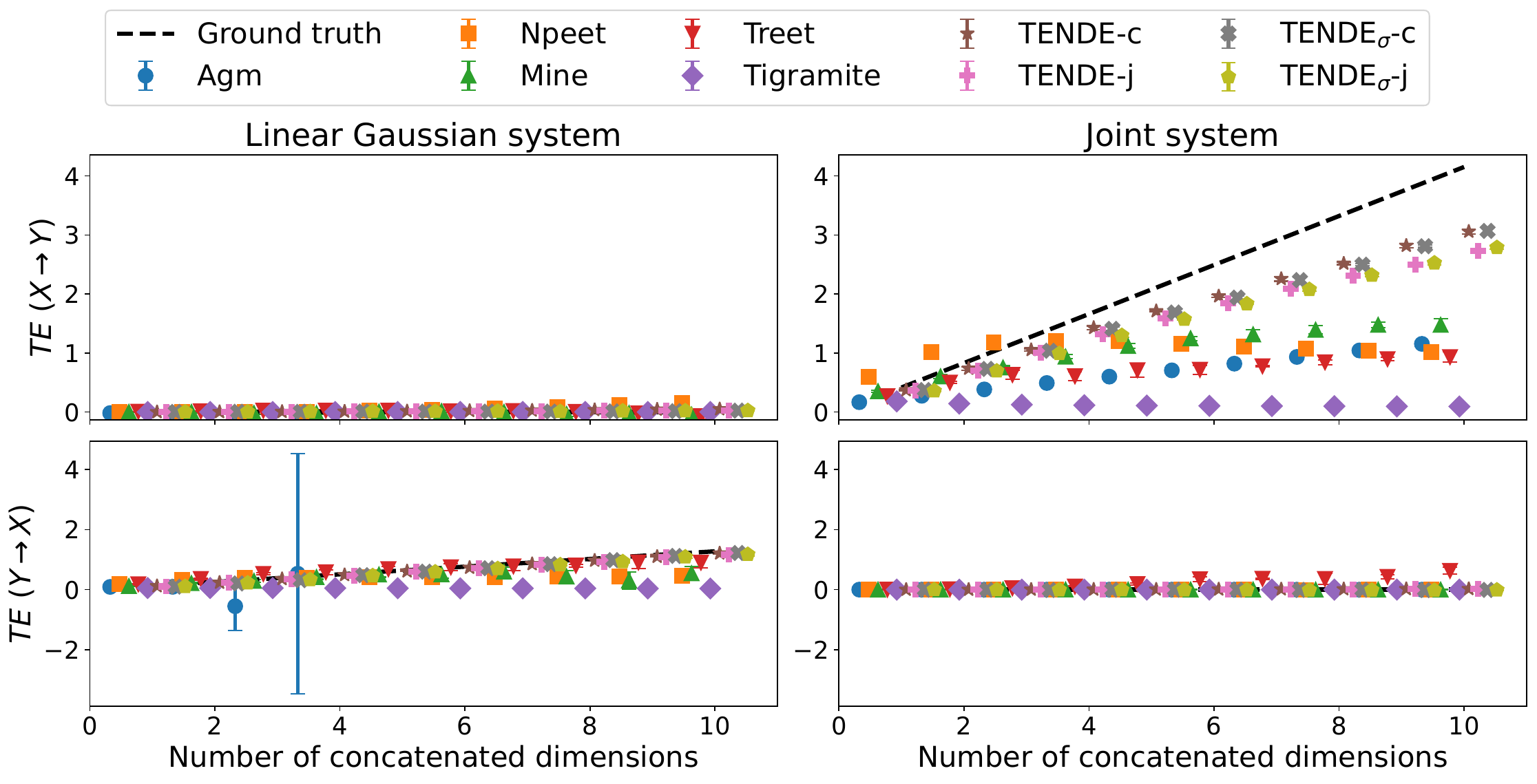}
    \includegraphics[width=0.48\textwidth]{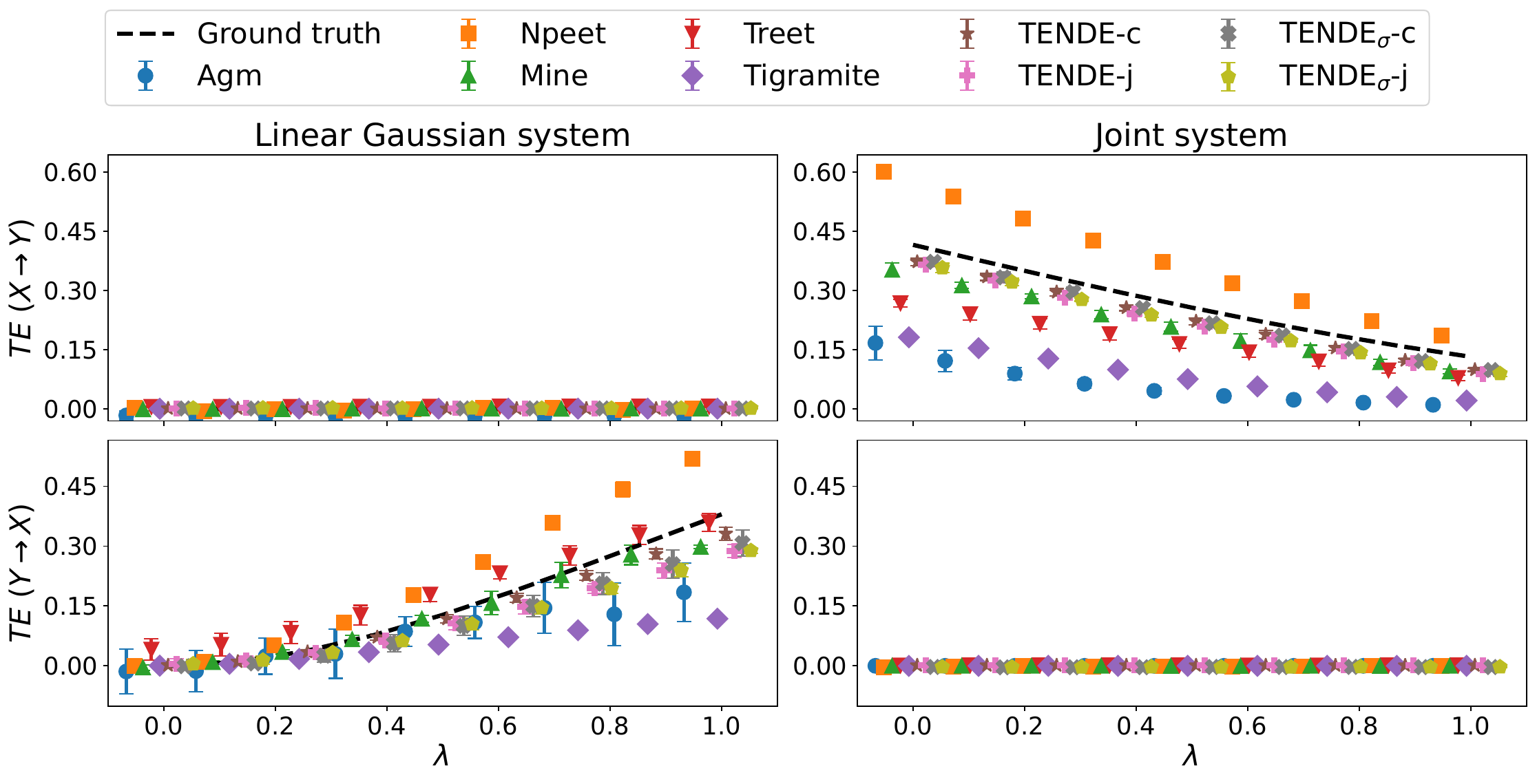}
    \caption{Estimated transfer entropy for the linear Gaussian and joint systems under redundant and linear stacking (top) and varying coupling strength $\lambda$ (bottom). Both systems are modified using the CDF mapping.}
    \label{fig:cdf}
\end{figure}
Across all configurations, the TENDE estimators continue to align closely with the analytical ground truth, confirming their robustness under the CDF transformation. In the redundant stacking scenario (top-left), TENDE correctly maintains stable estimates across varying numbers of redundant dimensions, while Treet exhibits large negative deviations with increasing variance and Tigramite yields near-zero estimates. In the linear stacking setting (top-right), where transfer entropy should increase linearly with the number of informative dimensions, TENDE maintains accurate scaling, whereas Treet collapses at higher dimensions and the remaining alternative methods consistently underestimate. For the simple coupling system (bottom), TENDE follows the expected monotonic trend with $\lambda$, closely matching the ground truth, while Npeet and Tigramite show noticeable deviations at stronger coupling values. 
\newpage

\subsection{Results at higher dimensions}
\label{sec:higher_dims}
We evaluate TENDE and the baseline estimators on higher-dimensional versions of the benchmarks described in \Secref{sec:estimation_benchmark}. In these experiments, the time series length is $T = 10000$ and both systems use 35 redundant or concatenated dimensions, resulting in 70-dimensional processes.
\subsubsection{Redundant stacking}
\label{sec:red_high}
\begin{table}[h!]
\centering
\caption{Estimated transfer entropy (mean $\pm$ standard deviation) versus ground truth for the 70-dimensional redundant stacking benchmark (linear Gaussian system, $T = 10000$). Results are sorted by proximity to the ground truth within each direction.}
\label{tab:redundant_70d_gaussian}
\small
\begin{tabular}{llllr}
\toprule
Direction & Method & Estimated TE $\pm$ Std & Ground Truth \\
\midrule
\multirow{8}{*}{$X \to Y$}
  & Mine          & $0.00 \pm 0.02$ & $0.0000$ \\
  & TENDE$_\sigma$-j & $0.05 \pm 0.00$ & $0.0000$ \\
  & TENDE-j       & $0.05 \pm 0.00$ & $0.0000$ \\
  & TENDE$_\sigma$-c & $0.06 \pm 0.01$ & $0.0000$ \\
  & TENDE-c       & $0.07 \pm 0.00$ & $0.0000$ \\
  & Agm           & $0.12 \pm 0.00$ & $0.0000$ \\
  & Npeet         & $-0.00 \pm 0.02$ & $0.0000$ \\
  & Treet         & $-3.02 \pm 2.59$ & $0.0000$ \\
\midrule
\multirow{8}{*}{$Y \to X$}
  & Mine          & $0.06 \pm 0.04$ & $0.1276$ \\
  & TENDE$_\sigma$-j & $0.20 \pm 0.02$ & $0.1276$ \\
  & TENDE-j       & $0.21 \pm 0.02$ & $0.1276$ \\
  & TENDE$_\sigma$-c & $0.23 \pm 0.02$ & $0.1276$ \\
  & Agm           & $0.24 \pm 0.01$ & $0.1276$ \\
  & TENDE-c       & $0.25 \pm 0.01$ & $0.1276$ \\
  & Npeet         & $0.00 \pm 0.01$ & $0.1276$ \\
  & Treet         & $-1.53 \pm 0.86$ & $0.1276$ \\
\bottomrule
\end{tabular}
\end{table}
\begin{table}[h!]
\centering
\caption{Estimated transfer entropy (mean $\pm$ standard deviation) versus ground truth for the 70-dimensional redundant stacking benchmark (joint system, $T = 10000$). Results are sorted by proximity to the ground truth within each direction.}
\label{tab:redundant_70d_joint}
\small
\begin{tabular}{llllr}
\toprule
Direction & Method & Estimated TE $\pm$ Std & Ground Truth \\
\midrule
\multirow{8}{*}{$X \to Y$}
  & TENDE-c       & $0.36 \pm 0.04$ & $0.4152$ \\
  & TENDE$_\sigma$-c & $0.36 \pm 0.05$ & $0.4152$ \\
  & Agm           & $0.31 \pm 0.00$ & $0.4152$ \\
  & TENDE-j       & $0.20 \pm 0.07$ & $0.4152$ \\
  & TENDE$_\sigma$-j & $0.20 \pm 0.08$ & $0.4152$ \\
  & Mine          & $0.15 \pm 0.03$ & $0.4152$ \\
  & Npeet         & $0.02 \pm 0.01$ & $0.4152$ \\
  & Treet         & $-3.52 \pm 1.14$ & $0.4152$ \\
\midrule
\multirow{8}{*}{$Y \to X$}
  & TENDE$_\sigma$-j & $0.04 \pm 0.00$ & $0.0000$ \\
  & TENDE-j       & $0.05 \pm 0.00$ & $0.0000$ \\
  & TENDE$_\sigma$-c & $0.05 \pm 0.00$ & $0.0000$ \\
  & TENDE-c       & $0.05 \pm 0.00$ & $0.0000$ \\
  & Agm           & $0.11 \pm 0.01$ & $0.0000$ \\
  & Npeet         & $-0.00 \pm 0.01$ & $0.0000$ \\
  & Mine          & $-0.02 \pm 0.02$ & $0.0000$ \\
  & Treet         & $-2.60 \pm 1.57$ & $0.0000$ \\
\bottomrule
\end{tabular}
\end{table}
In the redundant stacking setting at 70 dimensions, the TENDE variants consistently rank among the top estimators in both systems. For the linear Gaussian system (\Cref{tab:redundant_70d_gaussian}), all methods correctly identify the null transfer entropy in the $X \to Y$ direction, while in the $Y \to X$ direction, TENDE variants provide estimates closest to the ground truth alongside Agm. For the joint system (\Cref{tab:redundant_70d_joint}), TENDE-c and TENDE$_\sigma$-c achieve the best approximation of the non-zero transfer entropy in the $X \to Y$ direction, and all TENDE variants remain close to zero in the null $Y \to X$ direction. Across both systems, Npeet fails to detect the non-zero transfer entropy, Treet produces negative estimates with high variance, and Mine underestimates substantially. These results confirm that the score-based framework remains robust to irrelevant noise dimensions even when the score network must process over 100 input variables.
\subsubsection{Linear stacking}
\label{sec:linear_high}
\begin{table}[h!]
\centering
\caption{Estimated transfer entropy (mean $\pm$ standard deviation) versus ground truth for the 70-dimensional linear stacking benchmark (linear Gaussian system, $T = 10000$). Results are sorted by proximity to the ground truth within each direction.}
\label{tab:stacking_70d_gaussian}
\small
\begin{tabular}{llllr}
\toprule
Direction & Method & Estimated TE $\pm$ Std & Ground Truth \\
\midrule
\multirow{8}{*}{$X \to Y$}
  & Agm           & $0.13 \pm 0.00$ & $0.0000$ \\
  & TENDE$_\sigma$-j & $0.20 \pm 0.01$ & $0.0000$ \\
  & TENDE-j       & $0.21 \pm 0.01$ & $0.0000$ \\
  & TENDE$_\sigma$-c & $0.28 \pm 0.01$ & $0.0000$ \\
  & TENDE-c       & $0.39 \pm 0.03$ & $0.0000$ \\
  & Npeet         & $0.52 \pm 0.01$ & $0.0000$ \\
  & Mine          & $-0.05 \pm 0.05$ & $0.0000$ \\
  & Treet         & $-0.69 \pm 2.32$ & $0.0000$ \\
\midrule
\multirow{8}{*}{$Y \to X$}
  & Agm           & $4.45 \pm 0.01$ & $4.4660$ \\
  & TENDE$_\sigma$-c & $4.52 \pm 0.06$ & $4.4660$ \\
  & TENDE-j       & $4.35 \pm 0.06$ & $4.4660$ \\
  & TENDE$_\sigma$-j & $4.34 \pm 0.06$ & $4.4660$ \\
  & TENDE-c       & $4.86 \pm 0.02$ & $4.4660$ \\
  & Mine          & $0.65 \pm 0.38$ & $4.4660$ \\
  & Npeet         & $0.14 \pm 0.00$ & $4.4660$ \\
  & Treet         & $-1.80 \pm 1.79$ & $4.4660$ \\
\bottomrule
\end{tabular}
\end{table}
\begin{table}[h!]
\centering
\caption{Estimated transfer entropy (mean $\pm$ standard deviation) versus ground truth for the 70-dimensional linear stacking benchmark (joint system, $T = 10000$). Results are sorted by proximity to the ground truth within each direction.}
\label{tab:stacking_70d_joint}
\small
\begin{tabular}{llllr}
\toprule
Direction & Method & Estimated TE $\pm$ Std & Ground Truth \\
\midrule
\multirow{8}{*}{$X \to Y$}
  & TENDE-c       & $6.94 \pm 0.80$ & $14.5314$ \\
  & TENDE$_\sigma$-c & $6.86 \pm 0.79$ & $14.5314$ \\
  & Agm           & $6.79 \pm 0.02$ & $14.5314$ \\
  & TENDE$_\sigma$-j & $4.67 \pm 0.23$ & $14.5314$ \\
  & TENDE-j       & $4.65 \pm 0.22$ & $14.5314$ \\
  & Mine          & $0.92 \pm 0.07$ & $14.5314$ \\
  & Npeet         & $0.81 \pm 0.01$ & $14.5314$ \\
  & Treet         & $-0.63 \pm 1.42$ & $14.5314$ \\
\midrule
\multirow{8}{*}{$Y \to X$}
  & Npeet         & $0.01 \pm 0.01$ & $0.0000$ \\
  & TENDE$_\sigma$-j & $0.04 \pm 0.00$ & $0.0000$ \\
  & TENDE-j       & $0.04 \pm 0.00$ & $0.0000$ \\
  & TENDE$_\sigma$-c & $0.05 \pm 0.01$ & $0.0000$ \\
  & TENDE-c       & $0.05 \pm 0.01$ & $0.0000$ \\
  & Agm           & $0.11 \pm 0.00$ & $0.0000$ \\
  & Mine          & $-0.01 \pm 0.01$ & $0.0000$ \\
  & Treet         & $-1.10 \pm 1.51$ & $0.0000$ \\
\bottomrule
\end{tabular}
\end{table}
In the linear stacking setting at 70 dimensions, the transfer entropy grows additively with the number of independent process copies, resulting in large ground truth values that are particularly challenging to estimate. For the linear Gaussian system (\Cref{tab:stacking_70d_gaussian}), Agm achieves the closest estimate in the $Y \to X$ direction, followed closely by the TENDE variants, all of which recover the ground truth of $4.47$ nats within a margin of $0.15$ nats. In the joint system (\Cref{tab:stacking_70d_joint}), the ground truth of $14.53$ nats proves challenging for all methods; nevertheless, TENDE-c and TENDE$_\sigma$-c achieve the best approximations at approximately $6.9$ nats, outperforming Agm ($6.8$) and substantially outperforming Mine, Npeet, and Treet, which all remain below $1$ nat. In both systems, all TENDE variants correctly identify the null direction, and Treet consistently produces negative estimates with high variance. These results suggest that while the score-based framework scales better than competing approaches, very high-dimensional stacking settings with large ground truth values remain challenging and benefit from increased sample sizes, as explored in \Cref{tab:more_data}.
\begin{table}[h!]
\centering
\caption{Effect of increasing sample size on transfer entropy estimation for the 70-dimensional linear stacking benchmark (joint system). Results compare $T = 10000$ and $T = 50000$ observations, sorted by proximity to the ground truth at $T = 50000$.}
\label{tab:more_data}
\small
\begin{tabular}{lllrr}
\toprule
Direction & Method & Ground Truth & $T = 10000$ & $T = 50000$ \\
\midrule
\multirow{8}{*}{$X \to Y$}
  & TENDE-c       & $14.5314$ & $6.94 \pm 0.80$ & $10.86 \pm 0.10$ \\
  & TENDE$_\sigma$-c & $14.5314$ & $6.86 \pm 0.79$ & $10.83 \pm 0.09$ \\
  & TENDE-j       & $14.5314$ & $4.65 \pm 0.22$ & $10.35 \pm 0.15$ \\
  & TENDE$_\sigma$-j & $14.5314$ & $4.67 \pm 0.23$ & $10.33 \pm 0.15$ \\
  & Agm           & $14.5314$ & $6.79 \pm 0.02$ & $6.50 \pm 0.01$ \\
  & Mine          & $14.5314$ & $0.92 \pm 0.07$ & $1.24 \pm 0.08$ \\
  & Npeet         & $14.5314$ & $0.81 \pm 0.01$ & $1.00 \pm 0.00$ \\
  & Treet         & $14.5314$ & $-0.63 \pm 1.42$ & $-1.08 \pm 3.16$ \\
\midrule
\multirow{8}{*}{$Y \to X$}
  & TENDE$_\sigma$-c & $0.0000$ & $0.05 \pm 0.01$ & $0.01 \pm 0.00$ \\
  & TENDE-c       & $0.0000$ & $0.05 \pm 0.01$ & $0.01 \pm 0.00$ \\
  & TENDE-j       & $0.0000$ & $0.04 \pm 0.00$ & $0.01 \pm 0.00$ \\
  & TENDE$_\sigma$-j & $0.0000$ & $0.04 \pm 0.00$ & $0.01 \pm 0.00$ \\
  & Agm           & $0.0000$ & $0.11 \pm 0.00$ & $0.02 \pm 0.00$ \\
  & Mine          & $0.0000$ & $-0.01 \pm 0.01$ & $-0.00 \pm 0.00$ \\
  & Npeet         & $0.0000$ & $0.01 \pm 0.01$ & $-0.00 \pm 0.00$ \\
  & Treet         & $0.0000$ & $-1.10 \pm 1.51$ & $-1.49 \pm 0.86$ \\
\bottomrule
\end{tabular}
\end{table}
Increasing the sample size from $T = 10000$ to $T = 50000$ reveals a striking difference in how the estimators leverage additional data. In the $X \to Y$ direction, where the ground truth is $14.53$ nats, all four TENDE variants improve dramatically: TENDE-c rises from $6.94$ to $10.86$ nats, and even the joint variants more than double their estimates, while simultaneously reducing their standard deviations by an order of magnitude. In contrast, Agm shows no improvement (in fact slightly decreasing from $6.79$ to $6.50$), and Mine and Npeet remain below $1.3$ nats at both sample sizes. Treet worsens, producing more negative estimates with higher variance. In the null $Y \to X$ direction, all TENDE variants move closer to zero with more data, confirming that the improvements in the non-null direction are not artifacts of general overestimation. These results demonstrate that TENDE is uniquely capable of leveraging larger datasets in high-dimensional regimes, and suggest that with sufficient data the remaining gap to the ground truth can be further reduced.
\newpage
\section{Implementation details}
\label{sec:algos}
\paragraph{Unique denoising network.} For the implementation of TENDE, we adopt the Variance Preserving Stochastic Differential Equation framework \citep{song2020score}. The latter perturbs the data using an SDE parameterized by a drift $f_t$ and a diffusion coefficient $g_t$. Following \citet{e26040320}, we amortize the learning of all required parametric scores by using a single denoising score network.

\paragraph{Training.} Training is carried out through a randomized procedure. At each step, one of the possible encodings, which represents one of the score denoising score functions required for the computation of TE (joint, conditional, or marginal), is chosen. These denoising score functions are learned by the unique score network following the procedure described above. In total, estimating TE requires estimating either two or three score functions, which is something we achieve with a single denoising score network. 

\paragraph{SDE parameters.} We adopt the Variance Preserving framework proposed by \citet{song2020score}, where the drift and diffusion coefficients in \Eqref{eq:diffusion} are given by $f_t = -\frac{1}{2}\beta(t)$ and $g_t = \sqrt{\beta(t)}$, with a linear noise schedule $\beta(t) = \beta_{\min} + (\beta_{\max} - \beta_{\min})t$. In our implementation, we set $\beta_{\min} = 0.1$ and $\beta_{\max} = 20$. Under this parameterization, the transition density $p_{0t}(\cdot \vert x)$ is Gaussian with mean $k_t x$ and variance $\left(k_t^2 \int_0^t k_s^{-2} g_s^2 \, ds\right) \mathbf{I}_{N_\mathrm{x}}$, where $k_t = \exp\left\{\int_0^t f_s \, ds\right\}$, allowing exact sampling of $X_t \vert X_0 = x$ without numerical integration of the SDE.

\paragraph{Neural architecture, runtime, and preprocessing.} Our implementation adopts the architecture from the MINDE framework \citep{franzese2023minde}. The score network is a U-Net-style MLP with residual blocks, skip connections, and GroupNorm normalization. The network accepts as input the concatenation of the three variables $Y, X, Z$ as described in \Secref{sec:te_estimation_description}, along with an encoding vector $[e_1, e_2, e_3] \in \{-1, 0, 1\}^3$ that specifies the role of each input: $1$ indicates the variable for which the score is learned, $0$ denotes a conditioning signal (kept at its original value), and $-1$ indicates marginalization (the input is set to zero). The diffusion time $t^\star$ is embedded through a learned two-layer MLP with GELU activation and injected into each residual block via a scale-shift mechanism. The output layer is initialized to zero to ensure stable early training. We use the Adam optimizer with exponential moving average (momentum $m = 0.999$) and importance sampling at both training and inference time. For preprocessing, standard $z$-score normalization is applied to all time series prior to training. Regarding computational cost, the average runtime for estimating the transfer entropy between two one-dimensional time series with $T = 10000$ observations is approximately 20 minutes per pair of estimates (both directions) on a single NVIDIA A100 GPU.
\newpage
\begin{algorithm}[h!] 
\DontPrintSemicolon
\SetAlgoLined
\SetNoFillComment
\SetKwInOut{Parameter}{parameter}
\LinesNotNumbered 
\caption{TENDE (Single Training Step)}
\label{algo:tende_training}
\KwData{$ [X_t , Y_t, Z_t]$ }  
\Parameter{\texttt{approach} $\in \{\texttt{conditional}, \texttt{joint}\}$, $net_\theta( )$, with $\theta$ current parameters}
Obtain $Y, X, Z$ as described in \Secref{sec:te_estimation_description} \\
$t^\star \sim \mathcal{U}[0,T]$ \\
\tcp*{diffuse signals to timestep $t^\star$}
$ [Y_{t^\star}, X_{t^\star}, Z_{t^\star}] \gets k_{t^\star}[Y,X, Z]+\left(k^2_{t^\star}\int_0^{t^\star} k^{-2}_sg^2_{s}\dd s\right)^{\frac{1}{2}} [\epsilon_1,\epsilon_2, \epsilon_3]$, with $\epsilon_{1,2, 3} \sim \gamma_1$

\uIf{\texttt{approach} = \texttt{conditional}}{
    $c\sim \mathcal{U} \{0, 1\}$ \tcp*{Sample $c$ from a discrete uniform in $\{0, 1\}$}
}

\uElse{
    $c\sim \mathcal{U} \{0, 1, 2\}$ \tcp*{Sample $c$ from a discrete uniform in $\{0, 1, 2\}$}
}

\uIf{$c = 0 $ }{
\tcp*{Estimated conditional score on source and target} 
$\frac{\hat{\epsilon}}{\left(k^2_{t^\star}\int_0^{t^\star} k^{-2}_sg^2_{s}\dd s\right)^{\frac{1}{2}}} \gets net_\theta([Y_{t^\star} , X, Z], t^\star, [1, 0, 0] )$ 
}
\uElseIf{$c = 1$}{
\tcp*{Estimated conditional score only on the target} 
$\frac{\hat{\epsilon}}{\left(k^2_{t^\star}\int_0^{t^\star} k^{-2}_sg^2_{s}\dd s\right)^{\frac{1}{2}}}  \gets net_\theta([Y_{t^\star} , X, Z], t^\star, [1, -1, 0] ) $ 
}
\uElse{
\tcp*{Estimated unconditional score on the target}
$\frac{\hat{\epsilon}}{\left(k^2_{t^\star}\int_0^{t^\star} k^{-2}_sg^2_{s}\dd s\right)^{\frac{1}{2}}}  \gets net_\theta([Y_{t^\star} , X, Z], t^\star, [1, -1, -1] ) $ 
}
$L=\frac{g^2_{t^\star}}{\left(k^2_{t^\star}\int_0^{t^\star} k^{-2}_sg^2_{s}\dd s\right)^{\frac{1}{2}}}\norm{\epsilon-\hat{\epsilon}}^2$ \tcp*{ Compute Montecarlo sample associated to \Cref{eq:obj}} 
\Return Update $\theta$ according to gradient of $L$
\end{algorithm}

\end{document}